\documentclass[10pt,twocolumn,letterpaper]{article}
\IfFileExists{arxiv/arxiv_main.tex}{\def\SAGarxivpath{arxiv/}}{\def\SAGarxivpath{}}

\usepackage[pagenumbers]{\SAGarxivpath cvpr}

\usepackage{cuted} %

\usepackage{currfile} %

\usepackage{caption} %

\usepackage{microtype}

\usepackage{colortbl}
\usepackage{tabularx}
\usepackage{algorithm}
\usepackage{algpseudocode}
\usepackage{amsthm}

\newtheorem{lemma}{Lemma}[section]

\makeatletter
\renewcommand\paragraph{\@startsection{paragraph}{4}{\z@}
    {.5em}{-1em}{\normalfont\normalsize\bfseries}}
\makeatother

\usepackage{stfloats}

\definecolor{cvprblue}{rgb}{0.21,0.49,0.74}
\usepackage[pagebackref,breaklinks,colorlinks,allcolors=cvprblue]{hyperref}
\makeatletter
\providecommand{\theHALG@line}{}
\renewcommand{\theHALG@line}{\thealgorithm.\arabic{ALG@line}}
\makeatother
\hypersetup{
  pdftitle={Speculative Action Generation for World Action Models},
  pdfauthor={Lingkun Xiu, Xiaobo Wang, Bingwen Wei, Ruoyu Zhao, Jiahong Ning, Maoqing Li, Renquan Wan, Xiao-Ping Zhang, Jincheng Yu, Xinlei Chen}
}
\title{Speculative Action Generation for World Action Models}
\author{\parbox{0.96\textwidth}{\centering
\large
Lingkun Xiu$^{1}$\quad Xiaobo Wang$^{2,3,\dagger}$\quad Bingwen Wei$^{1}$\quad Ruoyu Zhao$^{4}$\quad Jiahong Ning$^{5}$\\[3pt]
Maoqing Li$^{6}$\quad Renquan Wan$^{6}$\quad Xiao-Ping Zhang$^{1}$\quad Jincheng Yu$^{1}$\quad Xinlei Chen$^{1,\dagger}$\\[7pt]
\normalsize
$^{1}$ Tsinghua University\quad $^{2}$ Shenzhen University of Advanced Technology\quad $^{3}$ Sangfor Technologies Inc.\\[3pt]
$^{4}$ City University of Hong Kong\quad $^{5}$ Infinigence AI\quad $^{6}$ AGIBot\\[5pt]
\small $^{\dagger}$Corresponding authors.
}}

\input{\SAGarxivpath title_layout}
\begin{document}
\maketitle
\raggedbottom
\suppressfloats[t]

\begin{abstract}
World Action Models (WAMs) have shown strong capabilities in robotic manipulation by jointly predicting actions and future visual observations.
However, iterative denoising incurs high inference latency, hindering continuous robot control in real time.
Although recent methods have reduced the cost of visual prediction at inference, action denoising still requires multiple sequential calls to the action diffusion transformer (DiT) and accounts for over 70\% of total inference latency.
To address this bottleneck, we propose \textbf{Speculative Action Generation (SAG)}, a plug-and-play speculative sampling framework that enables multiple denoising steps per action DiT call without additional training or auxiliary models.
Specifically, SAG implements two alternative algorithms: (1) \textbf{SAG-Residual} adapts speculative sampling to deterministic action denoising using a local residual criterion for acceptance and Euler updates for correction.
(2) \textbf{SAG-Coupling} reformulates action denoising as a stochastic differential equation to enable probabilistic acceptance and uses coupling for correction.
Extensive experiments on simulated and real-world tasks demonstrate that SAG accelerates WAM action generation by $1.7$--$3.3\times$ with comparable task performance and can be extended to flow-based vision language action (VLA) models with up to a $2.9\times$ speedup on $\pi_{0.5}$.
Our project page is available at \href{https://sag-wam.github.io/}{sag-wam.github.io}.
\end{abstract}

\begin{figure}[t]
    \centering
    \includegraphics[width=\linewidth]{\SAGarxivpath 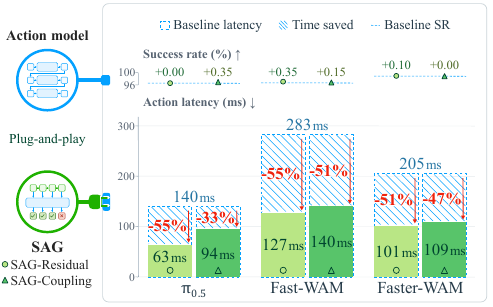}
    \caption{
    \textbf{SAG: A plug-and-play framework for accelerating action generation.}
    On LIBERO, both SAG variants reduce action generation latency by \textbf{33--55\%} across $\pi_{0.5}$, Fast-WAM, and Faster-WAM while maintaining task success rates.
    Solid bars show latency with SAG, and hatched regions indicate time saved relative to each baseline.
    Success rate annotations report changes from each baseline in percentage points.
    Best viewed in color.
    }
    \label{fig:sag_teaser}
\end{figure}

\section{Introduction}
\label{sec:intro}

World Action Models (WAMs) learn robot policies by jointly predicting actions and future visual observations~\cite{LiS-RSS-25,ZhuC-RSS-25,biMotusUnifiedLatent2025,kim2026cosmos,liCausalWorldModeling2026,ye2026worldactionmodelszeroshot}.
Mainstream WAMs couple a video diffusion transformer (DiT), which predicts future observations, with an action DiT conditioned on the resulting visual features~\cite{liCausalWorldModeling2026,zhouZeroWAMInContextWorldAction2026,huangLongWAMScalingContext2026,yuOpenWAMOpenFramework2026,zhangNativeVideoActionPretraining2026}.
However, both DiTs generate their outputs by iterative flow matching denoising and require multiple sequential calls, resulting in substantial inference latency~\cite{liCausalWorldModeling2026,huangLongWAMScalingContext2026,yuOpenWAMOpenFramework2026}.
Such latency can introduce delays between consecutive action chunks, thereby hindering WAM deployment for real-time, continuous robot control~\cite{huang2026streamingwam,akbariFlashWAMModalityAwareDistillation2026}.

Recent efforts have reduced this latency on the video side.
For instance, Fast-WAM retains video prediction as a training objective but skips future video generation at inference~\cite{yuanFastWAMWorldAction2026}.
Faster-WAM instead computes future visual features only once and reuses them at selected layers throughout action denoising~\cite{zhaoFasterWAMEfficientInferenceTime2026}.
However, with video latency reduced, sequential action denoising becomes the primary runtime bottleneck, consuming approximately 81--89\% of the total inference time on the LIBERO benchmark (Table~\ref{tab:libero_results}).
Reducing this latency without compromising task performance is challenging, since directly reducing the number of denoising steps can come at the cost of sample quality and each denoising step must wait for the output of the previous one~\cite{shihParallelSamplingDiffusion2023,bajpaiFlowCastTrajectoryForecasting2026}.
Step distillation shortens the denoising chain but requires additional training and may degrade the policy (Sec.~\ref{sec:related_work})~\cite{akbariFlashWAMModalityAwareDistillation2026,wangAnyStepWAMBudgetAlignedDistillation2026}.
We therefore aim to further accelerate action generation by reducing the number of sequential model calls per chunk, with no loss in task performance and no additional training.

Speculative sampling offers a natural way to reduce sequential model calls by exploiting parallel computation instead.
In large language models (LLMs), a faster draft model generates several draft tokens, and the target model verifies them in one parallel call without changing its output distribution~\cite{leviathanFastInferenceTransformers2023,chen2023acceleratinglargelanguagemodel}.
For action generation, draft states can be generated from the model's past velocity predictions without an auxiliary model, since the velocity field learned through flow matching is often locally smooth~\cite{pmlr-v267-deng25c,bajpaiFlowCastTrajectoryForecasting2026}.
The action DiT can then verify them in one parallel call.
However, deterministic sampling in flow matching prevents the direct application of the probabilistic acceptance and correction rules of standard speculative sampling~\cite{lipman2023flowmatchinggenerativemodeling,bortoliAcceleratedDiffusionModels2025}.

To this end, we propose \textbf{Speculative Action Generation (SAG)}, a plug-and-play acceleration framework for WAMs without additional training or auxiliary models.
SAG addresses this challenge through two alternative algorithms: \textbf{SAG-Residual} adapts the speculative sampling rules to the original action sampling process, whereas \textbf{SAG-Coupling} reformulates action sampling as a stochastic process to retain the original rules.
SAG-Residual extrapolates drafts along a recent velocity prediction, accepts drafts whose local Euler residuals do not exceed a tolerance, and corrects the first rejected draft with an Euler update, which bounds each step's deviation from an Euler step.
SAG-Coupling replaces the ordinary differential equation (ODE) for action generation with an equivalent stochastic differential equation (SDE), whose discretization yields Gaussian transition densities for probabilistic acceptance~\cite{song2021scorebasedgenerativemodelingstochastic}.
It drafts from a clean action estimate so that drafts respond to stochastic perturbations, and corrects the first rejected draft with reflection maximal coupling~\cite{bortoliAcceleratedDiffusionModels2025}, which preserves the distribution of the discrete stochastic target process.
SAG-Residual is faster in most of our experiments, whereas SAG-Coupling provides this distributional guarantee.

To evaluate SAG, we conduct extensive experiments with two representative WAMs, Fast-WAM and Faster-WAM, on three simulation benchmarks and four real robot tasks.
Across these settings, SAG reduces the number of action DiT calls per chunk from 10 to as few as 3, accelerating action generation by $1.7$--$3.3\times$ and overall inference by up to $2.6\times$, while keeping success rates within 1 percentage point of the original models in simulation (Fig.~\ref{fig:sag_teaser}).
It also complements step reduction, further accelerating sequential sampling at every tested number of denoising steps, and ablations confirm the role of each component.
Beyond WAMs, SAG can be extended to flow-based vision language action (VLA) models, achieving up to a $2.9\times$ speedup in action generation on $\pi_{0.5}$~\cite{intelligence2025pi05visionlanguageactionmodelopenworld}.
To sum up, our contributions~are:
\begin{itemize}
    \item We propose SAG, a plug-and-play speculative action generation framework for WAMs without additional training or auxiliary models.
    \item We develop two alternative algorithms: SAG-Residual adapts the speculative sampling rules to flow matching inference, while SAG-Coupling reformulates the inference process to retain the original rules.
    \item We demonstrate that SAG consistently accelerates action generation and maintains task performance for WAMs in simulation and real robot experiments.
    We further confirm that SAG can be extended to flow-based VLA models.
\end{itemize}

\begin{figure*}[!t]
    \centering
    \includegraphics[width=\textwidth]{\SAGarxivpath 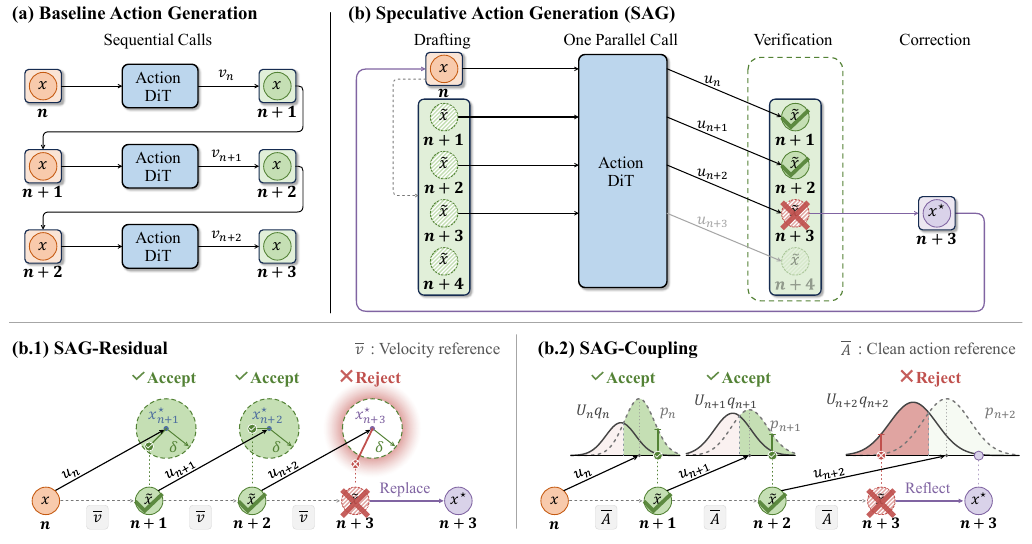}
    \caption{
    \textbf{Overview of Speculative Action Generation (SAG).}
    (a) The baseline generates actions through sequential action DiT calls, each advancing one denoising step.
    (b) SAG generates multiple drafts through simple mathematical computations and processes them in one parallel call, enabling multiple denoising steps per model call.
    (b.1) SAG-Residual generates draft states using a fixed velocity reference $\bar v$, accepts drafts whose local Euler residuals do not exceed $\delta$, and corrects the first rejected draft with an Euler update.
    (b.2) SAG-Coupling reconstructs the draft velocity from the clean action reference $\bar A$ to generate draft states, uses the log density ratio for probabilistic acceptance, and corrects the first rejected draft using reflection maximal coupling.
    }
    \label{fig:sag_overview}
\end{figure*}

\section{Related Work}
\label{sec:related_work}

\subsection{World Action Models}
Building on pretrained video models, WAMs jointly predict actions and future visual observations, demonstrating strong manipulation performance and generalization~\cite{biMotusUnifiedLatent2025,liCausalWorldModeling2026,ye2026worldactionmodelszeroshot,kim2026cosmos,shen2025videovlavideogeneratorsgeneralizable}.
Mainstream WAMs generate future observations with a video DiT and condition an action DiT on the resulting visual features~\cite{liCausalWorldModeling2026,zhouZeroWAMInContextWorldAction2026,huangLongWAMScalingContext2026,yuOpenWAMOpenFramework2026,zhangNativeVideoActionPretraining2026}.
However, both DiTs require multiple sequential calls to generate their outputs through iterative denoising, incurring high inference latency.
Recent methods seek to reduce video latency by simplifying visual computation~\cite{pai2025mimicvideovideoactionmodelsgeneralizable,ma2026dit4ditjointlymodelingvideo}.
For instance, Fast-WAM retains video prediction during training but skips future generation at inference~\cite{yuanFastWAMWorldAction2026}.
In contrast, Faster-WAM caches future visual features from a single video DiT pass and reuses them at selected action DiT layers throughout action denoising~\cite{zhaoFasterWAMEfficientInferenceTime2026}.
Both methods retain iterative action denoising, which becomes the primary runtime bottleneck.
Other methods shorten the denoising chain through distillation.
Flash-WAM distills video and action generation into one denoising step per modality using separate consistency functions~\cite{akbariFlashWAMModalityAwareDistillation2026}.
AnyStep-WAM further trains a scheduler to adaptively choose the number of denoising steps~\cite{wangAnyStepWAMBudgetAlignedDistillation2026}.
However, these distillation methods require additional training and may degrade model capabilities.
Our work instead reduces sequential action DiT calls without additional training while maintaining task performance.

\subsection{Speculative Sampling}
Speculative sampling is a technique for accelerating autoregressive decoding in LLMs without changing the target distribution~\cite{leviathanFastInferenceTransformers2023,chen2023acceleratinglargelanguagemodel,NEURIPS2023_6034a661,pmlr-v235-li24bt,zhangDraftVerifyLossless2024,xia-etal-2024-unlocking,caiMedusaSimpleLLM2024,chenSequoiaScalableRobust2024,liEAGLE3ScalingInference2025}.
A faster draft model proposes tokens that the target model verifies in parallel with probabilistic acceptance and correction~\cite{leviathanFastInferenceTransformers2023,chen2023acceleratinglargelanguagemodel,miaoSpecInferAcceleratingLarge2024,sunBlockVerificationAccelerates2025}.
For continuous diffusion states, speculative sampling can retain probabilistic acceptance and correct rejected drafts with reflection maximal coupling~\cite{bortoliAcceleratedDiffusionModels2025}.
FlowCast extends speculative sampling to flow matching for image and video generation by extrapolating intermediate denoising states and verifying them using a velocity error threshold~\cite{bajpaiFlowCastTrajectoryForecasting2026}.
For flow-based VLAs, FLASH trains a draft model to generate complete action chunks, which the target model checks in parallel to decide how many initial actions to execute~\cite{niu2026realtimevlaflashspeculativeinference}.
Unlike FLASH, we draft intermediate denoising states of WAMs from the target model's own predictions, without an auxiliary model.
Compared with FlowCast, SAG-Residual verifies drafts with step-scaled Euler residuals and reuses predictions across windows to avoid extra model calls.
SAG-Coupling instead brings coupling-based correction~\cite{bortoliAcceleratedDiffusionModels2025} to flow matching through an SDE reformulation and clean action drafting.
To our knowledge, this is the first application of speculative sampling to WAM action denoising.

\section{Preliminaries}
\label{sec:preliminaries}

\subsection{Flow Matching}
\label{sec:flow_matching}
The action DiT trained with flow matching generates action chunks by integrating a learned velocity field $v_\theta$ conditioned on $c$, which encodes observations and task information~\cite{lipman2023flowmatchinggenerativemodeling}.
With $c$ fixed, the action state $x(\tau)$ evolves from Gaussian noise at $\tau=1$ to an action chunk at $\tau=0$ according to the deterministic ODE:
\begin{equation}
    \mathrm dx(\tau)=v_\theta(x(\tau),\tau;c)\,\mathrm d\tau.
    \label{eq:flow_ode}
\end{equation}

At inference, we discretize Eq.~\eqref{eq:flow_ode} using forward Euler on the decreasing time grid $1=\tau_0>\cdots>\tau_K=0$, giving
\begin{equation}
    x_{k+1}=x_k-h_kv_k(x_k), x_0\sim\mathcal N(0,I).
    \label{eq:euler_update}
\end{equation}
Here, $x_k$ is the intermediate action state at $\tau_k$, $h_k=\tau_k-\tau_{k+1}>0$ is the step size, and $v_k(x)$ denotes $v_\theta(x,\tau_k;c)$.
Each velocity prediction depends on the state produced by the preceding update, so generating the final action chunk $x_K$ requires $K$ sequential action DiT calls.

\subsection{Speculative Sampling}
\label{sec:speculative_sampling}
Speculative sampling is a parallelizable inference strategy for accelerating autoregressive generation, particularly in LLMs~\cite{leviathanFastInferenceTransformers2023}.
For a token sequence $\mathbf y$, we denote its joint probability under the target model by $p(\mathbf y)$ and under the draft model by $q(\mathbf y)$.
Speculative sampling uses a computationally cheaper draft model to generate multiple candidates, which the target model then verifies in parallel with a single call.
Draft candidates are tested in order, and $\widetilde y_i$ at position $i$ is accepted if
\begin{equation}
    U_i\leq\alpha_i(\widetilde y_i)=\min\left\{1,\frac{p(\widetilde y_i\mid y_{<i})}{q(\widetilde y_i\mid y_{<i})}\right\},
    \label{eq:speculative_verify}
\end{equation}
where $y_{<i}$ denotes the prefix preceding position $i$, and $U_i\sim\operatorname{Unif}(0,1)$ is drawn independently for each candidate.

At the first rejection, one needs to sample a replacement from the adjusted distribution:
\begin{equation}
    y_i\sim\operatorname{norm}\!\left(\max\left(0,p(\cdot\mid y_{<i})-q(\cdot\mid y_{<i})\right)\right),
    \label{eq:speculative_correction}
\end{equation}
where $\operatorname{norm}$ denotes normalization.
With this replacement, the procedure discards all subsequent drafts and starts a new round of drafting from the corrected prefix.
If all candidates are accepted, drafting instead resumes from the prefix extended with the full draft sequence.
This procedure accelerates inference by reducing sequential target model calls without changing the target model's output distribution.
\par

\section{Speculative Action Generation}
\label{sec:method}

Typically, existing WAMs generate an action chunk through sequential denoising, with each action DiT call depending on the state produced by the preceding update.
To accelerate this process while maintaining action quality, we introduce two training-free algorithms based on speculative sampling (Sec.~\ref{sec:speculative_sampling}).
SAG-Residual adapts the speculative sampling rules to the original action sampling process, whereas SAG-Coupling reformulates this process to retain the original rules.
Both generate drafts at low computational cost without an auxiliary model and verify them in parallel, enabling multiple integration steps per target model call (Fig.~\ref{fig:sag_overview}).
From the current state $x_n$, we generate $L=\min(M,K-n)$ draft states, where $M$ is the maximum window length and $\widetilde x_k$ denotes a draft state.
We denote the predicted velocity $v_\theta(\widetilde x_k,\tau_k;c)$ by $u_k$.
Appendix~\ref{app:sampling_algorithms} of the supplementary material provides pseudocode for both SAG variants.

\subsection{SAG-Residual}
\label{sec:sag_residual}

To adapt speculative sampling to deterministic action generation, we replace the probabilistic acceptance rule with a local residual criterion and use Euler updates for correction.

\paragraph{Velocity drafting.}
During training, the model $v_\theta$ is supervised to match a constant target velocity along each linear interpolation path between noise and action samples.
This supervision encourages local smoothness of the velocity field in space and time.
At inference, we exploit this local smoothness to generate $L$ draft states from $x_n$ using a draft velocity $\bar v$ held fixed within each window:
\begin{equation}
    \widetilde x_k=x_n-(\tau_n-\tau_k)\bar v,
    k=n+1,\ldots,n+L.
    \label{eq:residual_draft}
\end{equation}
For the first window, we obtain $\bar v=v_0(x_0)$ from a single call to the target model, whereas subsequent windows reuse a prediction from the previous verification batch.

\paragraph{Verification.}
We obtain velocity predictions along the drafted trajectory in a single batched call to the target model.
Using these predictions, we compare each draft state $\widetilde x_{k+1}$ with the corresponding Euler update $x^\star_{k+1}=\widetilde x_k-h_ku_k$.
If the local Euler residual between these states does not exceed a predefined threshold $\delta\geq0$, we accept the draft, i.e.,
\begin{equation}
    R_k=\left\|\widetilde x_{k+1}-x^\star_{k+1}\right\|_2=h_k\|u_k-\bar v\|_2\leq\delta.
    \label{eq:residual_test}
\end{equation}
Here, $k=n,\ldots,n+L-1$ and $\widetilde x_n=x_n$.
Since $R_k$ scales the velocity error by $h_k$, a single $\delta$ bounds each accepted draft's deviation from its Euler update, even when step sizes vary.

\paragraph{Correction.}
We accept drafts in order and reject the first draft $\widetilde x_{j+1}$ with $R_j>\delta$.
We replace it with the corrected state $x_{j+1}=x^\star_{j+1}$ and discard all subsequent drafts.
We then restart drafting from $x_{j+1}$ with $\bar v=u_j$.
If all drafts are accepted, we instead advance to $x_{n+L}=\widetilde x_{n+L}$ and set $\bar v=u_{n+L-1}$ for the next window.

\subsection{SAG-Coupling}
\label{sec:sag_coupling}

To retain probabilistic acceptance and correction for action generation, we convert the action ODE into an SDE and apply reflection maximal coupling to its discretized transitions.

\paragraph{Action sampling reformulation.}
To apply speculative sampling to the action sampling process, we need a target transition density to compute the acceptance probability in Eq.~\eqref{eq:speculative_verify}.
To describe sampling with an increasing time variable, we set $r=1-\tau$ and denote $v_\theta(x_r,1-r;c)$ by $v_r$.
Using the relation between an SDE and its probability flow ODE~\cite{song2021scorebasedgenerativemodelingstochastic}, we convert the action ODE in Eq.~\eqref{eq:flow_ode} into an equivalent SDE by adding diffusion and correcting the drift to preserve the original time marginals when the velocity and score fields are exact,
\begin{equation}
    \mathrm d x_r\!=\!
    \underbrace{\bigl[-v_r+\tfrac{\eta^2(r)}{2}\nabla_x\log p_{1-r}(x_r)\bigr]}_{\text{Corrected drift}}
    \mathrm d r+
    \underbrace{\eta(r)\,\mathrm d B_r}_{\text{Diffusion term}},
    \label{eq:coupling_sampler_sde}
\end{equation}
where $\eta(r)$ is the noise scale, $\nabla_x\log p_{1-r}(x_r)=-[x_r+rv_r]/(1-r)$ is the score function of the marginal density $p_{1-r}$ (Appendix~\ref{app:score_identity}), with equality for the exact velocity field, and $B_r$ is a standard Brownian motion.

We apply the Euler--Maruyama method to discretize the SDE in Eq.~\eqref{eq:coupling_sampler_sde} for action sampling.
Over a step of size $h_k$ starting at $r_k=1-\tau_k$, the Brownian increment can be written as $\Delta B_k=B_{r_k+h_k}-B_{r_k}=\sqrt{h_k}\,z_k$, where $z_k\sim\mathcal N(0,I)$.
We choose $\eta(r)=\lambda\sqrt{1-r}$ with noise strength $\lambda>0$ to cancel the $1/(1-r)$ factor in the score correction.
By substituting this schedule and the score function into Eq.~\eqref{eq:coupling_sampler_sde} and using the Brownian increment above, we derive the final discrete update for action sampling:
\begin{equation}
    x_{k+1}=\underbrace{x_k-h_k\bigl[a_kv_k(x_k)+\tfrac{\lambda^2}{2}x_k\bigr]}_{m_k^P(x_k)}+\sigma_kz_k,
    \label{eq:coupling_target}
\end{equation}
where $a_k=1+\lambda^2r_k/2$ and $\sigma_k=\lambda\sqrt{h_k(1-r_k)}$.
Since $z_k$ is standard Gaussian, this update defines the target transition $P_k(\cdot\mid x_k)=\mathcal N(m_k^P(x_k),\sigma_k^2I)$, with density $p_k$.

\paragraph{Clean action drafting.}
To make drafts respond to the stochastic perturbations introduced by $\sigma_k z_k$ in Eq.~\eqref{eq:coupling_target}, we cache a clean action estimate $\widehat A_k(x)=x-\tau_kv_k(x)$.
For the first window, we initialize the reference $\bar A=\widehat A_0(x_0)$ with a single target model call, whereas subsequent windows reuse a clean action estimate from the previous verification batch.
We keep $\bar A$ fixed within each window and reconstruct the draft velocity as $\widetilde v_k=(\widetilde x_k-\bar A)/\tau_k$.
Thus, drafting involves only simple computations and requires no further calls to the target model.
Starting from $\widetilde x_n=x_n$, we generate $L$ draft states by replacing the model velocity in Eq.~\eqref{eq:coupling_target} with $\widetilde v_k$, retaining the same coefficients and noise scale:
\begin{equation}
    \widetilde x_{k+1}=\underbrace{\widetilde x_k-h_k\bigl[a_k\widetilde v_k+\tfrac{\lambda^2}{2}\widetilde x_k\bigr]}_{m_k^Q}+\sigma_kz_k.
    \label{eq:coupling_draft}
\end{equation}
This update defines the Gaussian draft transition ${Q_k(\cdot\mid\widetilde x_k)}=\mathcal N(m_k^Q,\sigma_k^2I)$, with density $q_k$ and the same covariance as $P_k$.

\begin{table*}[!b]
    \centering
    \caption{\textbf{Results on LIBERO.} We report success rates (\%), mean latencies (ms), and speedups for action generation and overall policy inference. Both SAG variants use $M=4$. SAG-Residual uses $\delta=0.05$ for the WAMs and $\delta=0.40$ for $\pi_{0.5}$. SAG-Coupling uses $\lambda=0.1$.}
    \label{tab:libero_results}
    \small
    \setlength{\tabcolsep}{1.6bp}
    \renewcommand{\arraystretch}{1.03}
    \begin{tabular}{>{\raggedright\arraybackslash}p{73.8000bp}*{5}{>{\centering\arraybackslash}p{42.4000bp}}>{\raggedleft\arraybackslash}p{44.8000bp}>{\raggedleft\arraybackslash}p{43.8000bp}>{\raggedleft\arraybackslash}p{44.8000bp}>{\raggedleft\arraybackslash}p{43.8000bp}}
    \toprule
        & \multicolumn{5}{c}{Success rate (\%) $\uparrow$} & \multicolumn{2}{c}{Action latency} & \multicolumn{2}{c}{Overall latency} \\
        \cmidrule(lr){2-6}\cmidrule(lr){7-8}\cmidrule(l){9-10}
        Method & \multicolumn{1}{c}{Spatial} & \multicolumn{1}{c}{Object} & \multicolumn{1}{c}{Goal} & \multicolumn{1}{c}{Long} & \multicolumn{1}{c}{Avg.} & \multicolumn{1}{r}{\makebox[28bp][c]{ms $\downarrow$}} & \multicolumn{1}{c}{\makebox[0pt][c]{Speedup $\uparrow$}} & \multicolumn{1}{r}{\makebox[28bp][c]{ms $\downarrow$}} & \multicolumn{1}{c}{\makebox[0pt][c]{Speedup $\uparrow$}} \\
        \midrule
        $\pi_{0.5}$ & 98.2 & 98.6 & 97.2 & 92.4 & \underline{96.60} & 140.04 & $1.00\times$ & 175.18 & $1.00\times$ \\
        \rowcolor{black!4}
        \hspace*{3bp}+ SAG-Residual & 99.2 & 98.0 & 97.0 & 92.2 & \underline{96.60} & 63.34 & $\mathbf{2.21\times}$ & 98.86 & $\mathbf{1.77\times}$ \\
        \rowcolor{black!4}
        \hspace*{3bp}+ SAG-Coupling & 98.6 & 97.8 & 96.8 & 94.6 & \textbf{96.95} & 94.40 & $\underline{1.48\times}$ & 129.72 & $\underline{1.35\times}$ \\
        \midrule
        Fast-WAM & 97.2 & 99.6 & 95.4 & 94.6 & 96.70 & 282.88 & $1.00\times$ & 317.36 & $1.00\times$ \\
        \rowcolor{black!4}
        \hspace*{3bp}+ SAG-Residual & 97.8 & 99.8 & 96.0 & 94.6 & \textbf{97.05} & 127.17 & $\mathbf{2.22\times}$ & 165.13 & $\mathbf{1.92\times}$ \\
        \rowcolor{black!4}
        \hspace*{3bp}+ SAG-Coupling & 97.6 & 99.4 & 96.2 & 94.2 & \underline{96.85} & 139.85 & $\underline{2.02\times}$ & 175.60 & $\underline{1.81\times}$ \\
        \midrule
        Faster-WAM & 99.4 & 99.6 & 98.2 & 98.2 & \underline{98.85} & 204.88 & $1.00\times$ & 251.44 & $1.00\times$ \\
        \rowcolor{black!4}
        \hspace*{3bp}+ SAG-Residual & 99.0 & 99.8 & 98.0 & 99.0 & \textbf{98.95} & 101.38 & $\mathbf{2.02\times}$ & 148.11 & $\mathbf{1.70\times}$ \\
        \rowcolor{black!4}
        \hspace*{3bp}+ SAG-Coupling & 98.6 & 99.8 & 99.2 & 97.8 & \underline{98.85} & 108.56 & $\underline{1.89\times}$ & 157.23 & $\underline{1.60\times}$ \\
        \bottomrule
    \end{tabular}
\end{table*}

\paragraph{Verification.}
We obtain velocity predictions along the drafted trajectory in a single batched call to the target model.
Each prediction $u_k$ gives the corresponding target mean $m_k^P=m_k^P(\widetilde x_k)$ via Eq.~\eqref{eq:coupling_target}.
Since $P_k$ and $Q_k$ are Gaussian with the same covariance, we compute the log density ratio at $\widetilde x_{k+1}$ as
\begin{equation}
    \ell_k=\log\frac{p_k(\widetilde x_{k+1})}{q_k(\widetilde x_{k+1})}
    =-z_k^\top\Delta_k-\tfrac12\|\Delta_k\|_2^2,
    \label{eq:coupling_log_ratio}
\end{equation}
where $\Delta_k=(m_k^Q-m_k^P)/\sigma_k$ is the normalized mean difference.
We verify each draft state using the speculative sampling acceptance rule in Eq.~\eqref{eq:speculative_verify}, accepting $\widetilde x_{k+1}$ when $\log U_k\leq\min(0,\ell_k)$, with independent draws $U_k\sim\operatorname{Unif}(0,1)$.
We apply this criterion in order and stop at the first rejection.

\paragraph{Correction.}
For the first rejected draft $\widetilde x_{j+1}$, we use reflection maximal coupling~\cite{bortoliAcceleratedDiffusionModels2025} to sample a replacement from the distribution in Eq.~\eqref{eq:speculative_correction}, with $p$ and $q$ replaced by the densities $p_j$ and $q_j$, by reflecting the noise $z_j$ used to generate $\widetilde x_{j+1}$,
\begin{equation}
    x_{j+1}^{\star}=m_j^P+\sigma_j\bigl[z_j-2d_j(d_j^\top z_j)\bigr],
    \label{eq:coupling_reflection}
\end{equation}
where $d_j=\Delta_j/\|\Delta_j\|_2$ is the unit vector in the direction of the mean difference.
We discard all subsequent drafts and restart drafting from $x_{j+1}=x_{j+1}^{\star}$ with $\bar A=\widehat A_{j+1}(\widetilde x_{j+1})$.
If all drafts are accepted, we instead advance to the final draft state $\widetilde x_{n+L}$ and reuse its clean action estimate as the next reference.

Together, these verification and correction steps ensure that SAG-Coupling preserves the joint distribution of trajectories generated by sequentially applying Eq.~\eqref{eq:coupling_target} (see Appendix~\ref{app:path_proof} for the proof).

\section{Experiments}
We evaluate SAG on Fast-WAM and Faster-WAM across simulation benchmarks and real robot experiments, measuring task success rates and inference speedups.
We further assess its applicability to the VLA model $\pi_{0.5}$~\cite{intelligence2025pi05visionlanguageactionmodelopenworld}.
We keep model weights fixed in all comparisons, and applying SAG requires no additional training.
Unless otherwise specified, all policies are evaluated with $K=10$ integration steps.
We measure action generation latency from action noise initialization to the output action chunk and overall inference latency from prepared inputs to the same output.
Both are measured using CUDA events on fixed inputs and averaged over ten runs after five warmup runs.

\subsection{Simulation Experiments}
\label{sec:simulation_experiments}
All simulation evaluations are conducted on two NVIDIA H100 GPUs for LIBERO and LIBERO-Plus and four NVIDIA A800 GPUs for RoboTwin~2.0.

\begin{table*}[!t]
    \centering
    \caption{\textbf{Results on LIBERO-Plus.} We report success rates (\%), mean latencies (ms), and speedups for action generation and overall policy inference. Avg. is computed as total successes divided by total trials. Both SAG variants use $M=4$. SAG-Residual uses $\delta=0.05$ for the WAMs and $\delta=0.40$ for $\pi_{0.5}$. SAG-Coupling uses $\lambda=0.1$.}
    \label{tab:libero_plus_results}
    \small
    \setlength{\tabcolsep}{1.6bp}
    \renewcommand{\arraystretch}{1.03}
    \begin{tabular}{>{\raggedright\arraybackslash}p{73.8000bp}*{8}{>{\centering\arraybackslash}p{25.3000bp}}>{\raggedleft\arraybackslash}p{44.8000bp}>{\raggedleft\arraybackslash}p{43.8000bp}>{\raggedleft\arraybackslash}p{44.8000bp}>{\raggedleft\arraybackslash}p{43.8000bp}}
    \toprule
        & \multicolumn{8}{c}{Success rate (\%) $\uparrow$} & \multicolumn{2}{c}{Action latency} & \multicolumn{2}{c}{Overall latency} \\
        \cmidrule(lr){2-9}\cmidrule(lr){10-11}\cmidrule(l){12-13}
        Method & \multicolumn{1}{c}{\makebox[0pt][c]{Cam.}} & \multicolumn{1}{c}{\makebox[0pt][c]{Robot}} & \multicolumn{1}{c}{\makebox[0pt][c]{Lang.}} & \multicolumn{1}{c}{\makebox[0pt][c]{Light}} & \multicolumn{1}{c}{\makebox[0pt][c]{Backg.}} & \multicolumn{1}{c}{\makebox[0pt][c]{Noise}} & \multicolumn{1}{c}{\makebox[0pt][c]{Layout}} & \multicolumn{1}{c}{\makebox[0pt][c]{Avg.}} & \multicolumn{1}{r}{\makebox[28bp][c]{ms $\downarrow$}} & \multicolumn{1}{c}{\makebox[0pt][c]{Speedup $\uparrow$}} & \multicolumn{1}{r}{\makebox[28bp][c]{ms $\downarrow$}} & \multicolumn{1}{c}{\makebox[0pt][c]{Speedup $\uparrow$}} \\
        \midrule
        $\pi_{0.5}$ & 65.83 & 82.32 & 86.48 & 96.30 & 94.59 & 84.05 & 85.45 & \underline{84.13} & 139.03 & $1.00\times$ & 174.41 & $1.00\times$ \\
        \rowcolor{black!4}
        \hspace*{3bp}+ SAG-Residual & 68.75 & 78.79 & 86.48 & 97.53 & 98.38 & 84.82 & 85.45 & \textbf{84.86} & 59.72 & $\mathbf{2.33\times}$ & 93.89 & $\mathbf{1.86\times}$ \\
        \rowcolor{black!4}
        \hspace*{3bp}+ SAG-Coupling & 65.83 & 80.30 & 87.70 & 98.77 & 94.05 & 83.27 & 85.45 & \underline{84.13} & 82.47 & $\underline{1.69\times}$ & 119.12 & $\underline{1.46\times}$ \\
        \midrule
        Fast-WAM & 16.25 & 41.92 & 72.13 & 84.57 & 47.57 & 33.07 & 60.00 & 49.14 & 280.83 & $1.00\times$ & 315.14 & $1.00\times$ \\
        \rowcolor{black!4}
        \hspace*{3bp}+ SAG-Residual & 17.08 & 44.44 & 72.95 & 83.33 & 48.11 & 33.85 & 58.64 & \underline{49.60} & 126.90 & $\mathbf{2.21\times}$ & 165.05 & $\mathbf{1.91\times}$ \\
        \rowcolor{black!4}
        \hspace*{3bp}+ SAG-Coupling & 17.08 & 44.95 & 73.77 & 82.72 & 46.49 & 32.68 & 62.73 & \textbf{49.93} & 141.20 & $\underline{1.99\times}$ & 177.17 & $\underline{1.78\times}$ \\
        \midrule
        Faster-WAM & 54.58 & 68.69 & 95.08 & 95.68 & 58.38 & 63.81 & 83.18 & \textbf{73.64} & 205.74 & $1.00\times$ & 260.53 & $1.00\times$ \\
        \rowcolor{black!4}
        \hspace*{3bp}+ SAG-Residual & 54.17 & 70.71 & 95.49 & 95.06 & 58.38 & 62.26 & 80.45 & 73.17 & 118.57 & $\underline{1.74\times}$ & 165.75 & $\underline{1.57\times}$ \\
        \rowcolor{black!4}
        \hspace*{3bp}+ SAG-Coupling & 53.75 & 69.19 & 95.49 & 94.44 & 58.92 & 63.81 & 83.18 & \underline{73.57} & 110.01 & $\mathbf{1.87\times}$ & 159.50 & $\mathbf{1.63\times}$ \\
        \bottomrule
    \end{tabular}
\end{table*}

\begin{table}[!b]
    \centering
    \caption{\textbf{Results on RoboTwin~2.0.} We report success rates (\%) and mean latencies (ms) for action generation and overall policy inference. SAG-Residual uses $M=8$, with $\delta=0.1$ for the WAMs and $\delta=0.40$ for $\pi_{0.5}$. SAG-Coupling uses $M=4$ and $\lambda=0.1$. At the final step, $\lambda$ is set to $0.01$.}
    \label{tab:robotwin_results}
    \small
    \setlength{\tabcolsep}{1.6bp}
    \renewcommand{\arraystretch}{1.03}
    \begin{tabular}{>{\raggedright\arraybackslash}p{73.8000bp}>{\raggedleft\arraybackslash}p{26.8000bp}>{\raggedleft\arraybackslash}p{23.8000bp}>{\raggedleft\arraybackslash}p{23.8000bp}>{\raggedleft\arraybackslash}p{35.8000bp}>{\raggedleft\arraybackslash}p{33.0500bp}}
        \toprule
        & \multicolumn{3}{c}{Success (\%) $\uparrow$} & \multicolumn{2}{c}{Latency (ms) $\downarrow$} \\
        \cmidrule(lr){2-4}\cmidrule(l){5-6}
        Method & \multicolumn{1}{r}{\makebox[22bp][c]{Clean}} & \multicolumn{1}{r}{\makebox[22bp][c]{Rand.}} & \multicolumn{1}{r}{\makebox[22bp][c]{Avg.}} & \multicolumn{1}{r}{\makebox[28bp][c]{Action}} & \multicolumn{1}{r}{\makebox[28bp][c]{Overall}} \\
        \midrule
        $\pi_{0.5}$ & 45.14 & 47.86 & \underline{46.50} & 186.14 & 242.54 \\
        \rowcolor{black!4}
        \hspace*{3bp}+ SAG-Residual & 51.18 & 56.22 & \textbf{53.70} & \textbf{65.16} & \textbf{122.25} \\
        \rowcolor{black!4}
        \hspace*{3bp}+ SAG-Coupling & 44.30 & 46.92 & 45.61 & \underline{133.68} & \underline{190.23} \\
        \midrule
        Fast-WAM & 90.84 & 89.98 & \textbf{90.41} & 428.95 & 481.69 \\
        \rowcolor{black!4}
        \hspace*{3bp}+ SAG-Residual & 90.76 & 89.98 & \underline{90.37} & \textbf{132.53} & \textbf{185.47} \\
        \rowcolor{black!4}
        \hspace*{3bp}+ SAG-Coupling & 90.42 & 89.08 & 89.75 & \underline{201.13} & \underline{251.51} \\
        \midrule
        Faster-WAM & 90.82 & 90.88 & \underline{90.85} & 286.41 & 346.86 \\
        \rowcolor{black!4}
        \hspace*{3bp}+ SAG-Residual & 91.36 & 91.22 & \textbf{91.29} & \textbf{87.52} & \textbf{148.19} \\
        \rowcolor{black!4}
        \hspace*{3bp}+ SAG-Coupling & 90.92 & 90.40 & 90.66 & \underline{120.37} & \underline{180.37} \\
        \bottomrule
    \end{tabular}
\end{table}

\subsubsection{Benchmarks}
\paragraph{LIBERO.}
LIBERO~\cite{liu2023liberobenchmarkingknowledgetransfer} is a robot simulation benchmark that consists of four distinct task suites: LIBERO-Spatial, LIBERO-Object, LIBERO-Goal, and LIBERO-Long, each containing 10 tasks.
We evaluate each task over 50 rollouts using fixed random seeds, totaling 2,000 rollouts.

\paragraph{LIBERO-Plus.}
LIBERO-Plus~\cite{fei2025liberoplusindepthrobustnessanalysis} extends LIBERO by introducing seven types of perturbations for robustness evaluation: camera viewpoints, robot initial states, language instructions, lighting, backgrounds, sensor noise, and object layouts.
We evaluate on a randomly sampled $15\%$ subset of the benchmark.

\paragraph{RoboTwin 2.0.}
RoboTwin~2.0~\cite{chen2025robotwin20scalabledata} is a simulation benchmark with 50 bimanual manipulation tasks.
We evaluate each task over 100 rollouts in each of the clean and randomized settings, totaling 10,000 rollouts.

\subsubsection{Main Results}
\paragraph{LIBERO.}
Both SAG variants substantially reduce action generation latency while maintaining comparable average success rates on LIBERO (Table~\ref{tab:libero_results}).
On Fast-WAM, SAG-Residual and SAG-Coupling achieve $2.22\times$ and $2.02\times$ speedups in action generation, respectively, with average success rates of $97.05\%$ and $96.85\%$, compared with $96.70\%$ for the baseline.
These gains translate into overall inference speedups of $1.92\times$ and $1.81\times$.
On Faster-WAM, which already simplifies visual computation, the two variants provide further action generation speedups of $2.02\times$ and $1.89\times$, with average success rates of $98.95\%$ and $98.85\%$, compared with $98.85\%$ for the baseline.
These benefits also extend to the VLA model $\pi_{0.5}$, with action generation speedups of $2.21\times$ and $1.48\times$ and comparable task performance.
To assess the impact of the SDE reformulation in SAG-Coupling on task performance, we compare the serial SDE and Euler samplers on Faster-WAM.
The SDE sampler performs comparably to the Euler sampler ($98.65\%$ vs.\ $98.85\%$ average success; Table~\ref{tab:app_libero_ode_sde}).

\paragraph{LIBERO-Plus.}
On LIBERO-Plus, both SAG variants achieve aggregate success rates within $0.8$ percentage points of the corresponding baselines across all three models (Table~\ref{tab:libero_plus_results}).
SAG-Residual achieves action generation speedups of $1.74\times$ to $2.33\times$, and SAG-Coupling achieves $1.69\times$ to $1.99\times$.

\paragraph{RoboTwin 2.0.}
RoboTwin experiments use two wrist cameras and one head camera for bimanual manipulation, processing one additional camera view compared with LIBERO and thus incurring higher inference latency.
SAG-Residual achieves higher action generation speedups of $3.24\times$ on Fast-WAM and $3.27\times$ on Faster-WAM while maintaining task performance (Table~\ref{tab:robotwin_results}).
This gain partly reflects SAG-Residual's larger window on RoboTwin ($M=8$).
SAG-Coupling achieves action generation speedups of $2.13\times$ on Fast-WAM and $2.38\times$ on Faster-WAM, with success rates close to the corresponding baselines.
Its speedup increase over LIBERO is smaller than that of SAG-Residual.
We reduce $\lambda$ from $0.1$ to $0.01$ in the final step to preserve task success on RoboTwin, which may limit further acceleration.
On $\pi_{0.5}$, SAG-Residual also achieves a $2.86\times$ action generation speedup, while SAG-Coupling achieves a $1.39\times$ speedup.

\paragraph{Sources of acceleration.}
For Faster-WAM on RoboTwin with \mbox{$K=10$}, serial sampling requires 10 action DiT calls per chunk, whereas SAG-Residual requires as few as 3 and \mbox{SAG-Coupling~4}.
The measured action generation speedups ($3.27\times$ and $2.38\times$) approach these best-case ratios ($3.3\times$ and $2.5\times$), so batched verification adds little per-call latency at these window sizes.

Since SAG leaves visual processing and variational autoencoder (VAE) encoding unchanged, its overall speedup is smaller than its action generation speedup.

\subsubsection{Effect of Sampling Steps}
\label{sec:sampling_steps}
SAG can adapt to different numbers of integration steps, allowing users to choose the sampling budget according to their needs.
To demonstrate this adaptability, we compare serial Euler sampling with both SAG variants at $K\in\{5,10,20\}$ on LIBERO-Spatial for $\pi_{0.5}$, Fast-WAM, and Faster-WAM.

\begin{table}[!ht]
    \centering
    \caption{\textbf{Effect of Sampling Steps on LIBERO-Spatial.} We report success rates (SR, \%) and mean action generation latencies (Action, ms) for $K\in\{5,10,20\}$ integration steps. Other SAG parameters follow Table~\ref{tab:libero_results}.}
    \label{tab:sampling_steps}
    \fontsize{8.5}{10.5}\selectfont
    \setlength{\tabcolsep}{1.6bp}
    \renewcommand{\arraystretch}{1.08}
    \begin{tabular}{>{\raggedright\arraybackslash}p{61.3000bp}>{\raggedleft\arraybackslash}p{9.0500bp}>{\raggedleft\arraybackslash}p{19.3000bp}>{\raggedleft\arraybackslash}p{26.8000bp}>{\raggedleft\arraybackslash}p{20.3000bp}>{\raggedleft\arraybackslash}p{26.8000bp}>{\raggedleft\arraybackslash}p{20.3000bp}>{\raggedleft\arraybackslash}p{26.8000bp}}
        \toprule
        & & \multicolumn{2}{c}{$\pi_{0.5}$} & \multicolumn{2}{c}{Fast-WAM} & \multicolumn{2}{c}{Faster-WAM} \\
        \cmidrule(lr){3-4}\cmidrule(lr){5-6}\cmidrule(l){7-8}
        Method & $K$ & \multicolumn{1}{r}{\makebox[14.9bp][c]{SR}} & \multicolumn{1}{r}{\makebox[23.3bp][c]{Action}} & \multicolumn{1}{r}{\makebox[14.9bp][c]{SR}} & \multicolumn{1}{r}{\makebox[23.3bp][c]{Action}} & \multicolumn{1}{r}{\makebox[14.9bp][c]{SR}} & \multicolumn{1}{r}{\makebox[23.3bp][c]{Action}} \\
        \midrule
         & 20 & 99.0 & 294.86 & 97.2 & 567.59 & 98.8 & 396.87 \\
        Serial Euler & 10 & 98.2 & 150.82 & 97.2 & 280.30 & 99.4 & 205.33 \\
         & 5 & 97.8 & 80.97 & 97.0 & 138.33 & 98.4 & 99.87 \\
        \midrule
        \rowcolor{black!4}
         & 20 & 99.2 & \textbf{101.22} & 97.4 & \textbf{198.19} & 98.6 & \textbf{185.38} \\
        \rowcolor{black!4}
        \hspace*{3bp}+ SAG-Residual & 10 & 99.2 & \textbf{69.53} & 97.8 & \textbf{125.30} & 99.0 & \underline{135.32} \\
        \rowcolor{black!4}
         & 5 & 98.4 & \textbf{50.23} & 97.2 & \textbf{86.59} & 98.6 & \underline{81.48} \\
        \midrule
        \rowcolor{black!4}
         & 20 & 98.0 & \underline{187.82} & 97.6 & \underline{204.99} & 99.0 & \underline{201.21} \\
        \rowcolor{black!4}
        \hspace*{3bp}+ SAG-Coupling & 10 & 98.6 & \underline{105.91} & 97.6 & \underline{132.60} & 98.6 & \textbf{120.81} \\
        \rowcolor{black!4}
         & 5 & 98.4 & \underline{71.99} & 97.6 & \underline{99.19} & 99.2 & \textbf{72.51} \\
        \bottomrule
    \end{tabular}
\end{table}

Compared with serial Euler sampling at the same $K$, both SAG variants reduce action generation latency for all three models at every tested $K$, with success rate differences of at most 1 percentage point (Table~\ref{tab:sampling_steps}).
We report speedups relative to each model's serial Euler baseline at the same $K$.
At $K=20$, SAG-Residual achieves $2.86\times$ and $2.14\times$ speedups on Fast-WAM and Faster-WAM, respectively, while SAG-Coupling achieves $2.77\times$ and $1.97\times$.
At $K=10$, SAG-Residual achieves $2.24\times$ and $1.52\times$ speedups, while SAG-Coupling achieves $2.11\times$ and $1.70\times$.
Even at $K=5$, SAG-Residual still achieves $1.60\times$ and $1.23\times$ speedups, while SAG-Coupling achieves $1.39\times$ and $1.38\times$.
Both variants require one target model call to initialize the drafting reference.
This fixed cost is harder to amortize at smaller $K$, contributing to lower speedups.
SAG thus complements step reduction: with $K=5$ on both WAMs, SAG generates actions $4.9$--$6.6\times$ faster than serial sampling with $K=20$, with success rates within 0.4 percentage points.

\begin{figure}[!t]
    \centering
    \includegraphics[width=\columnwidth]{\SAGarxivpath 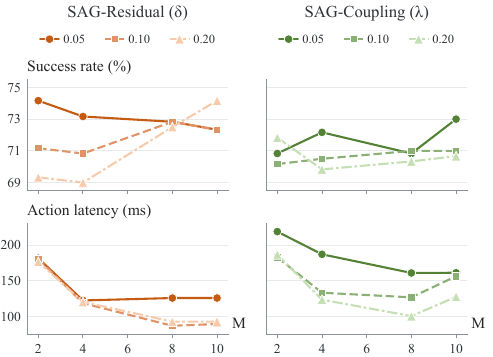}
    \caption{\textbf{Ablation Study of SAG Hyperparameters.}
    All experiments are conducted in the randomized setting.
    We vary $M$ and $\delta$ for SAG-Residual and $M$ and $\lambda$ for SAG-Coupling, using Faster-WAM with $K=10$.
    For SAG-Coupling, $\lambda$ is set to $0.01$ at the final step.}
    \label{fig:parameter_sensitivity}
\end{figure}

\subsubsection{Ablation Studies}
We conduct ablation studies on Faster-WAM to investigate the impact of SAG hyperparameters and core components.
For these studies, we select six tasks from RoboTwin~2.0 with substantial variation in denoising velocity, excluding tasks with high baseline success rates to reduce ceiling effects (see Appendix~\ref{app:robotwin_subset_studies} for details).

\paragraph{Hyperparameter ablations.}
We investigate the influence of window length $M$ on both SAG variants, residual threshold $\delta$ on SAG-Residual, and noise strength $\lambda$ on SAG-Coupling, as shown in Fig.~\ref{fig:parameter_sensitivity}.
For SAG-Residual, we observe that increasing $M$ from $2$ to $8$ reduces action generation latency by allowing more integration steps per model call.
However, latency rises slightly at $M=10$, as verifying more draft states incurs additional costs without necessarily advancing more integration steps.
For the residual threshold $\delta$, increasing it from $0.05$ to $0.1$ further reduces latency by allowing more draft states to pass verification, particularly for larger windows.
However, a larger threshold of $0.2$ offers little or no further acceleration and reduces success rates in most tested settings, as it permits larger deviations from the target Euler updates.

For SAG-Coupling, increasing $M$ from $2$ to $4$ substantially reduces action generation latency.
Further increasing $M$ to $8$ offers limited additional acceleration, but latency rises at $M=10$, consistent with the growing cost of parallel verification.
For the noise strength $\lambda$, increasing it from $0.05$ to $0.1$ can reduce action generation latency.
Larger noise increases the overlap between the target and draft transitions, so more drafts pass verification.
This acceleration comes with a success rate decrease of at most 2 percentage points, so we use $\lambda=0.1$ as the default trade-off.
At $\lambda=0.2$, latency decreases further at larger windows, but success rates decline in most tested settings.

\begin{table}[!t]
    \centering
    \captionof{table}{\textbf{Ablation Study of SAG Components.}
We evaluate Faster-WAM on the six selected RoboTwin tasks in the randomized setting, with 100 rollouts per task.
SAG parameters follow Table~\ref{tab:robotwin_results}.}
\label{tab:mechanism_ablation}
\begingroup
\small
\setlength{\tabcolsep}{0pt}
\renewcommand{\arraystretch}{1.08}
\noindent
\begin{tabular*}{\linewidth}{@{\extracolsep{\fill}}lrrr@{}}
\toprule
Variant & SR (\%) $\uparrow$ & Action (ms) $\downarrow$ & Overall (ms) $\downarrow$ \\
\midrule
\multicolumn{4}{@{}l@{}}{\textbf{(a) SAG-Residual}} \\
\addlinespace[2pt]
Full method & \textbf{72.83} & \textbf{87.17} & \textbf{147.85} \\
w/o residual check & 68.83 & 88.38 & 150.52 \\
w/o Euler correction & 70.50 & 87.53 & 149.98 \\
\midrule
\multicolumn{4}{@{}l@{}}{\textbf{(b) SAG-Coupling}} \\
\addlinespace[2pt]
Full method & \textbf{70.50} & \textbf{133.31} & \textbf{199.06} \\
Constant-velocity draft & 69.17 & 240.77 & 302.17 \\
w/o reference reuse & 69.83 & 183.33 & 244.48 \\
\bottomrule
\end{tabular*}
\par
\endgroup

\end{table}

\paragraph{Component ablations.}
We further evaluate the contributions of the core components in both SAG variants (Table~\ref{tab:mechanism_ablation}).
For SAG-Residual, removing the residual check and Euler correction individually reduces success rates by $4.00$ and $2.33$ percentage points, respectively, while action generation latency remains nearly unchanged.
This suggests that both components help maintain task performance by limiting deviations from the target Euler updates.

For SAG-Coupling, replacing clean action drafting with constant velocity drafting increases action generation latency by $80.6\%$, and removing reference reuse increases it by $37.5\%$.
These results indicate that adapting draft velocities to the current noisy state and reusing predictions from previous verification batches both contribute to efficient action generation.

\subsection{Real-World Experiments}
\subsubsection{Setup and Tasks}
\begin{figure}[!t]
    \centering
    \includegraphics[width=\linewidth]{\SAGarxivpath 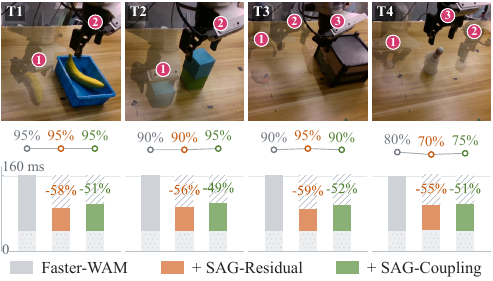}
    \caption{\textbf{Results on Real Robot Tasks.}
    Tasks are (T1) banana pick-and-place, (T2) block stacking, (T3) toy storage and drawer closing, and (T4) shuttlecock stacking.
    Upper lines show success rates (\%) and bars show inference latency (ms).
    Dotted bases show remaining latency, including VAE encoding and visual processing, while solid segments show action latency.
    Full bar heights represent overall latency.
    Hatched extensions show action time saved relative to Faster-WAM.
    Percentages inside the hatching report the action latency reduction.
    Image numbers indicate execution order.}
    \label{fig:real_world}
\end{figure}

We further evaluate the performance of SAG on four tabletop manipulation tasks in the real world: (T1) a banana pick-and-place task, (T2) a block stacking task, (T3) a toy storage task, and (T4) a shuttlecock stacking task.
We use a Franka single-arm robot with a Robotiq 2F-85 gripper.
Visual observations comprise a wrist RGB view and a third-person RGB view of the workspace.
For each task, we collect 50 demonstrations via VR-based teleoperation using a PICO headset and its handheld controllers.
The state and action spaces are consistent with the configuration in the LIBERO benchmark.
We finetune a Faster-WAM~\cite{zhaoFasterWAMEfficientInferenceTime2026} checkpoint on each task for 30 epochs using four NVIDIA A800 GPUs, with a batch size of 8 per GPU.
For each task, we deploy the finetuned model on an NVIDIA RTX 4090 GPU and evaluate the success rate and inference latency over 20 trials.
We show snapshots of these task executions in Fig.~\ref{fig:real_world}, numbered in execution order.
Tasks T1 and T2 involve grasping and placement, while T3 additionally requires closing the drawer after placing the toy inside.
The three snapshots for T4 illustrate successive stages of shuttlecock stacking. Additional task details are provided in Appendix~\ref{app:real_world_details}.

\subsubsection{Results}
We present the quantitative results in Fig.~\ref{fig:real_world}.
Compared with the Faster-WAM baseline, both SAG variants achieve comparable average success rates and reduce inference latency on all four real-world tasks.
On average, SAG-Residual and SAG-Coupling accelerate action generation by $2.33\times$ and $2.03\times$, respectively, with corresponding overall inference speedups of $1.71\times$ and $1.59\times$.
SAG-Residual achieves lower action generation latency because it advances an average of $4.00$ integration steps per full window, compared with $3.42$ for SAG-Coupling (Table~\ref{tab:app_real_world_sampler_profile}, Appendix~\ref{app:real_world_details}).
These results show that the speedups of SAG carry over to real-world manipulation.

\section{Conclusion}
\label{sec:conclusion}
We introduced SAG, a plug-and-play speculative sampling framework for accelerating action generation in WAMs without additional training or auxiliary models.
SAG drafts multiple intermediate action states from previous model predictions and verifies them in parallel, reducing sequential model calls.
Specifically, SAG-Residual adapts the speculative sampling rules to deterministic action sampling, whereas SAG-Coupling reformulates action sampling as a stochastic process to retain probabilistic verification and correction.
Extensive experiments in simulation and on real robots demonstrate that SAG reduces inference latency while maintaining task performance.
Results on $\pi_{0.5}$ further support its applicability to VLAs.

\section*{LLM Usage Statement}
We used OpenAI Codex to assist with writing and polishing this manuscript.
It also assisted with writing project code and plotting code.
The authors reviewed and verified all LLM-assisted text and code and take full responsibility for the content of the manuscript.

\par
{
    \small
    \bibliographystyle{\SAGarxivpath ieeenat_fullname}
    \bibliography{\SAGarxivpath main}
}
\backrefsetup{disable}
\clearpage
\appendix
\twocolumn[\centering\Large
  \vspace*{12.6935pt}
  \textbf{Appendix}\\
  \vspace{\dimexpr 1em+12.6935pt\relax}
]
\setcounter{equation}{0}
\renewcommand{\theequation}{\thesection.\arabic{equation}}
\numberwithin{equation}{section}
\raggedbottom
\setlength{\textfloatsep}{\intextsep}
\definecolor{sagstagecomment}{RGB}{0,128,0}
\newcommand{\sagstage}[2][\algorithmicindent]{%
    \Statex\hspace{#1}\textcolor{sagstagecomment}{$\triangleright$\ #2}%
}
\section{Sampling Algorithms}
\label{app:sampling_algorithms}

We follow the notation of Sec.~\ref{sec:method}, with $v_k(x)=v_\theta(x,\tau_k;c)$ and $h_k=\tau_k-\tau_{k+1}$.

\subsection{SAG-Residual}
Algorithm~\ref{alg:sag_residual} summarizes the sampling procedure for SAG-Residual, including velocity drafting, batched verification, acceptance based on the local residual criterion, and Euler correction.

\begin{algorithm}[!htbp]
\caption{SAG-Residual}
\label{alg:sag_residual}
\small
\begin{algorithmic}[1]
\Require Target model $v_\theta$, condition $c$, initial state $x_0$, grid $\{\tau_k\}_{k=0}^K$, $M\geq1$, $\delta\geq0$
\Ensure Final action chunk $x_K$
\sagstage[0pt]{Initialization}
\State $n\gets0$; $\bar v\gets v_0(x_0)$ \Comment{One model call}
\While{$n<K$}
    \sagstage{Drafting}
    \State $L\gets\min(M,K-n)$; $\widetilde x_n\gets x_n$
    \For{$k=n,\ldots,n+L-1$}
        \State $\widetilde x_{k+1}\gets\widetilde x_k-h_k\bar v$
    \EndFor
    \sagstage{Batched verification}
    \State $(u_n,\ldots,u_{n+L-1})\gets$ \Comment{One batch call}
    \Statex \hspace{\algorithmicindent}\hspace{\algorithmicindent}$\bigl(v_n(\widetilde x_n),\ldots,v_{n+L-1}(\widetilde x_{n+L-1})\bigr)$
    \State Compute the following for $k=n,\ldots,n+L-1$:
    \Statex \hspace{\algorithmicindent}$x^\star_{k+1}\gets\widetilde x_k-h_ku_k$, $R_k\gets\|\widetilde x_{k+1}-x^\star_{k+1}\|_2$.
    \sagstage{Acceptance and correction}
    \State Accept drafts in order while $R_k\leq\delta$.
    \State $j\gets$ first rejected index, or $n+L$ if none
    \If{$j<n+L$}
        \State $x_{n+1:j}\gets\widetilde x_{n+1:j}$
        \State $x_{j+1}\gets x^\star_{j+1}$ \Comment{Discard later drafts}
        \State $\bar v\gets u_j$; $n\gets j+1$
    \Else
        \State $x_{n+1:n+L}\gets\widetilde x_{n+1:n+L}$
        \State $\bar v\gets u_{n+L-1}$; $n\gets n+L$
    \EndIf
\EndWhile
\State \Return $x_K$
\end{algorithmic}
\end{algorithm}

We reuse a velocity prediction from the verification batch to update $\bar v$ for the next window, without an additional model call.

\subsection{SAG-Coupling}
Algorithm~\ref{alg:sag_coupling} summarizes the sampling procedure for SAG-Coupling, including clean action drafting, batched verification, probabilistic acceptance, and correction via reflection maximal coupling.
We use the coefficients $a_k$ and $\sigma_k$ defined in Eq.~\eqref{eq:coupling_target}.

\begin{algorithm}[!htbp]
\caption{SAG-Coupling}
\label{alg:sag_coupling}
\small
\begin{algorithmic}[1]
\Require Target model $v_\theta$, condition $c$, initial state $x_0$, grid $\{\tau_k\}_{k=0}^K$, $M\geq1$, $\lambda>0$
\Ensure Final action chunk $x_K$
\sagstage[0pt]{Initialization}
\State $n\gets0$; $u_0\gets v_0(x_0)$ \Comment{One model call}
\State $\bar A\gets x_0-\tau_0u_0$
\While{$n<K$}
    \sagstage{Drafting}
    \State $L\gets\min(M,K-n)$; $\widetilde x_n\gets x_n$
    \For{$k=n,\ldots,n+L-1$}
        \State $\widetilde v_k\gets(\widetilde x_k-\bar A)/\tau_k$
        \State Draw fresh $z_k\sim\mathcal N(0,I)$.
        \State $m_k^Q\gets\widetilde x_k-h_k[a_k\widetilde v_k+\lambda^2\widetilde x_k/2]$
        \State $\widetilde x_{k+1}\gets m_k^Q+\sigma_kz_k$
    \EndFor
    \sagstage{Batched verification}
    \State $\mathcal I\gets\{n,\ldots,\min(n+L,K-1)\}$, excluding
    \Statex \hspace{\algorithmicindent}$n$ if $u_n$ is already available at $x_n$.
    \State If $\mathcal I\ne\varnothing$: \Comment{One batch call}
    \Statex \hspace{\algorithmicindent}\hspace{\algorithmicindent}$(u_k)_{k\in\mathcal I}\gets\bigl(v_k(\widetilde x_k)\bigr)_{k\in\mathcal I}$
    \State Compute the following for $k=n,\ldots,n+L-1$:
    \Statex \hspace{\algorithmicindent}$m_k^P\gets\widetilde x_k-h_k[a_ku_k+\lambda^2\widetilde x_k/2]$,
    \Statex \hspace{\algorithmicindent}$\Delta_k\gets(m_k^Q-m_k^P)/\sigma_k$; $\ell_k\gets-z_k^\top\Delta_k-\tfrac12\|\Delta_k\|_2^2$.
    \sagstage{Acceptance and correction}
    \State Accept drafts in order while $\log U_k\leq\min(0,\ell_k)$,
    \Statex \hspace{\algorithmicindent}with independent $U_k\sim\operatorname{Unif}(0,1)$.
    \State $j\gets$ first rejected index, or $n+L$ if none
    \If{$j<n+L$}
        \State $x_{n+1:j}\gets\widetilde x_{n+1:j}$; $d_j\gets\Delta_j/\|\Delta_j\|_2$
        \State $x_{j+1}\gets m_j^P+\sigma_j[z_j-2d_j(d_j^\top z_j)]$
        \State $m\gets j+1$ \Comment{Discard later drafts}
    \Else
        \State $x_{n+1:n+L}\gets\widetilde x_{n+1:n+L}$; $m\gets n+L$
    \EndIf
    \sagstage{Reference update}
    \If{$m<K$}
        \State $\bar A\gets\widetilde x_m-\tau_mu_m$
        \State Retain $u_m$ after full acceptance and invalidate
        \Statex \hspace{\algorithmicindent}\hspace{\algorithmicindent}it after rejection.
    \EndIf
    \State $n\gets m$
\EndWhile
\State \Return $x_K$
\end{algorithmic}
\end{algorithm}

The verification batch includes the nonterminal draft endpoint to update $\bar A$.
In the first window and after full acceptance, we use the target velocity already evaluated at $x_n$ as the first draft velocity.
After rejection, $\bar A$ is computed from the rejected draft, while sampling resumes from the corrected state.
The velocity at that corrected state is evaluated in the next verification batch.

\section{From the Action ODE to a Stochastic Target}
\label{app:ode_sde}

\newtheorem{sagbproposition}{Proposition}[section]

\subsection{Reformulating the Action ODE as an SDE}
\label{app:score_identity}

This section derives the stochastic reformulation in Sec.~\ref{sec:sag_coupling}, starting from the action ODE in Eq.~\eqref{eq:flow_ode}.
We first express the score required for the drift correction through the exact flow matching velocity, then construct the SDE in Eq.~\eqref{eq:coupling_sampler_sde} and prove preservation of the ODE's time marginals.
Fix the conditioning $c$ and suppress it throughout this subsection.
We flatten each action chunk into $\mathbb R^d$, where $d=HD$ for a chunk of shape $H\times D$.

\begin{lemma}[Score from the flow matching velocity]
\label{lem:app_score_velocity}
Let $A\sim\mu$ be a clean action chunk satisfying $\mathbb E\|A\|_2^2<\infty$, and let $\varepsilon\sim\mathcal N(0,I_d)$ be independent of $A$.
For the linear interpolation
\begin{equation}
    X_\tau=(1-\tau)A+\tau\varepsilon,
    \label{eq:flow_path}
\end{equation}
let $p_\tau$ denote its density for $0<\tau\leq1$ and define the exact velocity
\begin{equation}
    v^\star(x,\tau)
    =\mathbb E[\varepsilon-A\mid X_\tau=x].
    \label{eq:app_exact_flow_velocity}
\end{equation}
The score, the gradient of the log density with respect to $x$, satisfies
\begin{equation}
    \nabla_x\log p_\tau(x)
    =-\frac{x+(1-\tau)v^\star(x,\tau)}{\tau}.
    \label{eq:app_score}
\end{equation}
\end{lemma}

\begin{proof}
Conditioned on $A=a$, the interpolation is Gaussian with mean $(1-\tau)a$ and covariance $\tau^2I_d$.
Writing $\varphi_\tau$ for the density of $\mathcal N(0,\tau^2I_d)$ gives
\begin{equation}
    p_\tau(x)
    =\int\varphi_\tau\!\left(x-(1-\tau)a\right)\mu(\mathrm da).
    \label{eq:app_gaussian_mixture}
\end{equation}
For $\tau>0$, this density is strictly positive and smooth in $x$.
The Gaussian kernel bounds and the finite moment assumption justify spatial and time differentiation under the integral on intervals bounded away from $\tau=0$.
Differentiating with respect to $x$ and dividing by $p_\tau(x)$ yields
\begin{align}
    \nabla_x\log p_\tau(x)
    &=-\frac{1}{\tau^2}\mathbb E[x-(1-\tau)A\mid X_\tau=x]\notag\\
    &=-\frac1\tau\mathbb E[\varepsilon\mid X_\tau=x].
    \label{eq:app_score_conditional_noise}
\end{align}
Finally, the identity $\varepsilon=X_\tau+(1-\tau)(\varepsilon-A)$ and Eq.~\eqref{eq:app_exact_flow_velocity} give
\begin{equation*}
    \mathbb E[\varepsilon\mid X_\tau=x]
    =x+(1-\tau)v^\star(x,\tau),
\end{equation*}
which proves Eq.~\eqref{eq:app_score}.
\end{proof}

Setting $\tau=1-r$ recovers the score expression stated after Eq.~\eqref{eq:coupling_sampler_sde}.
With this expression available, we next determine the drift correction that allows diffusion to be added while preserving the ODE's marginals.
Let $\eta(r)\geq0$ be a deterministic continuous noise scale on $[0,1)$.

\begin{sagbproposition}[Marginal preservation]
\label{prop:app_marginal_preservation}
Consider the exact action ODE $\mathrm dx_r=-v^\star(x_r,1-r)\,\mathrm dr$ and the SDE
\begin{equation}
    \mathrm dx_r=b^\star(x_r,r)\,\mathrm dr+\eta(r)\,\mathrm dB_r,
    \label{eq:app_candidate_sde}
\end{equation}
with corrected drift
\begin{equation}
    b^\star(x,r)
    =-v^\star(x,1-r)
      +\frac{\eta(r)^2}{2}
       \underbrace{\nabla_x\log p_{1-r}(x)}_{\text{Score}}.
    \label{eq:app_velocity_sde_drift}
\end{equation}
Both processes start from $\mathcal N(0,I_d)$, with the SDE's initial state independent of future Brownian increments.
On every $[0,T]$ with $T<1$, assume that both processes admit nonexplosive solutions.
For each process, assume that its forward equation has a unique solution with this initial law among weakly continuous probability laws with an integrable expected drift norm.
Assume that the corresponding process marginals belong to this class.
Then both processes have density $p_{1-r}$ for $0\leq r<1$, and these common marginals converge weakly to $\mu$ as $r\uparrow1$.
\end{sagbproposition}

\begin{proof}
We first identify the density evolution to be preserved.
The conditional Gaussian path given $A=a$ has velocity $(x-a)/\tau$.
Averaging its continuity equation over $\mu$ gives $\partial_\tau p_\tau=-\nabla_x\cdot(p_\tau v^\star)$, using Eq.~\eqref{eq:app_exact_flow_velocity}, as in the flow matching construction~\cite{lipman2023flowmatchinggenerativemodeling}.
Under $r=1-\tau$, this becomes
\begin{equation}
    \partial_r p_{1-r}
    =\nabla_x\cdot\left(p_{1-r}v^\star(x,1-r)\right),
    \label{eq:app_continuity}
\end{equation}
the forward equation of the exact action ODE.

Adding diffusion while retaining the drift $-v^\star$ introduces an additional term in the density evolution.
Specifically, the SDE with drift $b(x,r)$ has the Fokker--Planck equation~\cite{song2021scorebasedgenerativemodelingstochastic}
\begin{equation}
    \partial_r\rho_r
    =-\nabla_x\cdot\left(b(\cdot,r)\rho_r\right)
      +\frac{\eta(r)^2}{2}\Delta_x\rho_r.
    \label{eq:app_fokker_planck}
\end{equation}
We therefore need a drift correction whose contribution cancels the diffusion term when $\rho_r=p_{1-r}$.
One sufficient choice is Eq.~\eqref{eq:app_velocity_sde_drift}, since $p_{1-r}\nabla_x\log p_{1-r}=\nabla_xp_{1-r}$ gives
\begin{align*}
    &-\nabla_x\cdot\left(b^\star(\cdot,r)p_{1-r}\right)
       +\frac{\eta(r)^2}{2}\Delta_xp_{1-r}\\
    &\qquad=\nabla_x\cdot\left(p_{1-r}v^\star(x,1-r)\right).
\end{align*}
Thus, the candidate density satisfies both forward equations.

To apply uniqueness, we verify the required integrability under the candidate marginals.
Conditional Jensen's inequality in Eqs.~\eqref{eq:app_exact_flow_velocity} and~\eqref{eq:app_score_conditional_noise} gives
\begin{align*}
    \mathbb E\|v^\star(X_\tau,\tau)\|_2
    &\leq\mathbb E\|\varepsilon-A\|_2,\\
    \mathbb E\|\nabla_x\log p_\tau(X_\tau)\|_2
    &\leq\frac{\mathbb E\|\varepsilon\|_2}{\tau}.
\end{align*}
Consequently,
\begin{equation*}
    \mathbb E\|b^\star(X_{1-r},r)\|_2
    \leq\mathbb E\|\varepsilon-A\|_2
      +\frac{\eta(r)^2}{2(1-r)}\mathbb E\|\varepsilon\|_2.
\end{equation*}
These bounds are uniform on each $[0,T]$, $T<1$, since $\eta$ is bounded there and $1-r\geq1-T$.
The interpolation laws are weakly continuous and have initial distribution $\mathcal N(0,I_d)$.
The uniqueness assumptions therefore imply that the ODE and SDE marginals equal $p_{1-r}$ on $[0,T]$.
Since $T<1$ is arbitrary, this proves marginal preservation for all $r<1$ and establishes Eq.~\eqref{eq:coupling_sampler_sde} for the exact field.

It remains to identify the terminal marginal.
Independence and $\mathbb E\varepsilon=0$ imply
\begin{equation}
    \mathbb E\|X_\tau-A\|_2^2
    =\tau^2\left(d+\mathbb E\|A\|_2^2\right)
    \xrightarrow[\tau\downarrow0]{}0.
    \label{eq:app_terminal_marginal}
\end{equation}
Hence $X_\tau$ converges to $A$ in mean square, and the common ODE and SDE marginals converge weakly to $\mu$.
\end{proof}

This result establishes the continuous reformulation for the exact velocity field.
We next derive the discrete target used in sampling with the learned model.

\subsection{Discretization of the SDE}
\label{app:discretization}

On the time grid of the main text, set $r_k=1-\tau_k$, with $0=r_0<\cdots<r_K=1$ and $h_k=r_{k+1}-r_k>0$.
For each $k<K$, prescribe a noise strength $\lambda_k>0$ and set $\eta(r_k)=\lambda_k\sqrt{1-r_k}$.

To construct a score approximation, we replace $v^\star(x,\tau_k)$ in Lemma~\ref{lem:app_score_velocity} with the learned velocity $v_k(x)=v_\theta(x,\tau_k;c)$.
Substituting this approximation and the prescribed noise scale into Eq.~\eqref{eq:app_velocity_sde_drift} cancels the explicit $1/(1-r_k)$ factor, giving the drift at the grid point:
\begin{equation}
    b_\theta(x,r_k)
    =-v_k(x)-\frac{\lambda_k^2}{2}\left[x+r_kv_k(x)\right].
    \label{eq:app_learned_drift}
\end{equation}

Euler--Maruyama evaluates both coefficients at $r_k$ and uses $\Delta B_k=\sqrt{h_k}\,z_k$, with independent $z_k\sim\mathcal N(0,I_d)$ also independent of $x_0$.
The drift increment $h_kb_\theta(x_k,r_k)$ and noise increment $\eta(r_k)\Delta B_k$ yield
\begin{equation}
    x_{k+1}=\underbrace{x_k-h_k\bigl[a_kv_k(x_k)+\tfrac{\lambda_k^2}{2}x_k\bigr]}_{m_k^P(x_k)}+\sigma_kz_k,
    \label{eq:app_target_parameters}
\end{equation}
where $a_k=1+\lambda_k^2r_k/2$ and $\sigma_k=\lambda_k\sqrt{h_k(1-r_k)}$.
Since $x_k$ depends only on $x_0$ and earlier increments, $z_k$ is independent of $x_k$.
For a fixed deterministic model, the conditional update therefore has the Gaussian law
\begin{equation}
    P_k(\cdot\mid x)
    =\mathcal N\!\left(m_k^P(x),\sigma_k^2I_d\right).
    \label{eq:app_discrete_kernel}
\end{equation}
When $\lambda_k\equiv\lambda$, this recovers Eq.~\eqref{eq:coupling_target}.
The default RoboTwin schedule uses $\lambda_k=0.1$ for $k<K-1$ and \mbox{$\lambda_{K-1}=0.01$}.

For every $k<K$, $h_k>0$ and $\tau_k>0$, so $\sigma_k^2>0$.
This includes the final transition to $\tau_K=0$, whose coefficients are evaluated at $\tau_{K-1}>0$.
No score or velocity evaluation at the clean endpoint is required, and the final transition still has a nondegenerate Gaussian density.

The Gaussian kernel is exact for the defined discrete update, which approximates the continuous dynamics.
The score inferred from $v_\theta$ need not equal the score of the learned ODE's own marginals, so Proposition~\ref{prop:app_marginal_preservation} does not imply equality between this finite-step chain and the original Euler sampler.
This discrete Gaussian chain is the target of SAG-Coupling.
Appendix~\ref{app:speculative_proof} proves that speculative acceptance and correction preserve its joint law under the stated model and arithmetic assumptions.

\section{Speculative Sampling on the Stochastic Action Chain}
\label{app:speculative_proof}

\newtheorem{sagcproposition}{Proposition}[section]

We prove the joint-distribution statement for SAG-Coupling in Sec.~\ref{sec:sag_coupling}.
We specialize the Gaussian reflection maximal coupling of De Bortoli et al.~\cite{bortoliAcceleratedDiffusionModels2025} to our transitions and show that parallel verification and reference reuse preserve the joint law of the discrete target chain.
Finally, we derive the conditional acceptance probability and its dependence on the reference.

\subsection{Conditional draft and target densities}
\label{app:conditional_densities}

Fix the conditioning $c$, the learned model, the time grid, and the prescribed noise schedule defining $P_k$ in Eq.~\eqref{eq:app_discrete_kernel}.
The strengths $\lambda_k>0$ may vary across steps, with $\lambda_k\equiv\lambda$ recovering Eq.~\eqref{eq:coupling_target}.
The target and draft use the same $\sigma_k>0$ at each step.
We assume exact arithmetic and a deterministic, statewise model: at a fixed state and time, individual and batched evaluations return the same velocity.

At logical step $k$, let $\mathcal H_k$ denote the information generated by the confirmed states $x_{0:k}$, the active reference $\bar A$, and the proposal and acceptance variables revealed before this transition.
The actual mean $m_k^Q$ used to generate the current draft is measurable with respect to $\mathcal H_k$.
The current draws $z_k\sim\mathcal N(0,I_d)$ and $U_k\sim\mathrm{Unif}(0,1)$ are independent of this history and of each other.
This history describes a logical reveal of random variables; it excludes unused future draws even if the implementation computes the window in advance.
Section~\ref{app:path_proof} verifies that the algorithm admits this representation.

Write $m_k^P=m_k^P(x_k)$ for the target mean at the confirmed predecessor.
Conditioned on $\mathcal H_k$, the draft and target densities are
\begin{equation}
    \begin{aligned}
        q_k(y\mid\mathcal H_k)&=\mathcal N(y;m_k^Q,\sigma_k^2I_d),\\
        p_k(y\mid x_k)&=\mathcal N(y;m_k^P,\sigma_k^2I_d).
    \end{aligned}
    \label{eq:app_conditional_densities}
\end{equation}
For $\widetilde x_{k+1}=m_k^Q+\sigma_kz_k$ and $\Delta_k=(m_k^Q-m_k^P)/\sigma_k$, the standardized residuals are
\begin{equation*}
    \frac{\widetilde x_{k+1}-m_k^Q}{\sigma_k}=z_k,
    \frac{\widetilde x_{k+1}-m_k^P}{\sigma_k}=z_k+\Delta_k.
\end{equation*}
The common Gaussian normalizer $(2\pi\sigma_k^2)^{-d/2}$ cancels in the density ratio, leaving
\begin{equation}
    \begin{aligned}
        \ell_k
        &=\log\frac{p_k(\widetilde x_{k+1}\mid x_k)}
                        {q_k(\widetilde x_{k+1}\mid\mathcal H_k)}\\
        &=-\frac{\|\widetilde x_{k+1}-m_k^P\|_2^2
                    -\|\widetilde x_{k+1}-m_k^Q\|_2^2}{2\sigma_k^2}\\
        &=-\frac12\|z_k+\Delta_k\|_2^2+\frac12\|z_k\|_2^2\\
        &=-z_k^\top\Delta_k-\frac12\|\Delta_k\|_2^2.
    \end{aligned}
    \label{eq:app_log_ratio_derivation}
\end{equation}
Thus Eq.~\eqref{eq:speculative_verify} accepts the candidate exactly when
\begin{equation}
    \log U_k\leq\min\{0,\ell_k\},
    \label{eq:coupling_ratio}
\end{equation}
Here $\ell_k$ is exactly the log density ratio in Eq.~\eqref{eq:coupling_log_ratio} of the main text.
The denominator must use the mean that actually generated this candidate, including when the reference was obtained from an earlier rejected draft.
Updating the reference affects only subsequent windows and cannot change the proposal density used to generate an existing candidate.

\subsection{Correctness and maximality}
\label{app:reflection_proof}

\begin{lemma}[Gaussian acceptance and reflection]
\label{lem:reflection}
Let $q$ and $p$ be the densities of $\mathcal N(m^Q,\sigma^2I_d)$ and $\mathcal N(m^P,\sigma^2I_d)$, respectively, with $\sigma>0$.
Draw independent $Z\sim\mathcal N(0,I_d)$ and $U\sim\mathrm{Unif}(0,1)$, and set $\widetilde Y=m^Q+\sigma Z$.
If $U\leq\min\{1,p(\widetilde Y)/q(\widetilde Y)\}$, output $Y=\widetilde Y$.
Otherwise, output $Y=m^P+\sigma(I_d-2ee^\top)Z$, where $e=(m^Q-m^P)/\|m^Q-m^P\|_2$.
Then $Y$ has density $p$.
When $m^Q=m^P$, acceptance is unconditional and $e$ need not be defined.
\end{lemma}

\begin{proof}
If the means coincide, $p=q$ and the accepted proposal already has the target density.
Suppose henceforth that $m^Q\ne m^P$.

\emph{Reflection and density exchange.}
Set $R=I_d-2ee^\top$ and define
\begin{equation}
    T(y)=m^P+R(y-m^Q).
    \label{eq:app_affine_reflection}
\end{equation}
Since $e^\top e=1$, we have
\begin{equation*}
    R^\top=R,
    R^2=I_d-4ee^\top+4ee^\top ee^\top=I_d.
\end{equation*}
Thus $R$ is orthogonal, $|\det R|=1$, and $R(m^Q-m^P)=-(m^Q-m^P)$.
Applying $T$ twice gives
\begin{equation*}
    T(T(y))=y-(I_d+R)(m^Q-m^P)=y.
\end{equation*}
Consequently, $T^{-1}=T$ and $|\det DT|=1$.
The differences from the two means satisfy
\begin{align*}
    T(y)-m^P&=R(y-m^Q),\\
    T(y)-m^Q&=R(y-m^P).
\end{align*}
Orthogonality preserves their squared norms, while the common covariance gives the same normalizing constant in both densities.
Therefore,
\begin{equation}
    p(T(y))=q(y), q(T(y))=p(y).
    \label{eq:app_density_swap}
\end{equation}

\emph{Probability mass in the two branches.}
Because $U$ is independent of the candidate, the accepted branch contributes the unnormalized output density
\begin{equation*}
    a(y)=q(y)\min\{1,p(y)/q(y)\}=\min\{q(y),p(y)\}.
\end{equation*}
The unnormalized density of rejected candidates is
\begin{equation*}
    r(y)=q(y)-a(y)=[q(y)-p(y)]_+,
\end{equation*}
where $[b]_+=\max(b,0)$.
On rejection, Eq.~\eqref{eq:app_affine_reflection} maps the candidate to $T(\widetilde Y)=m^P+\sigma RZ$, exactly the correction in the lemma.
The change-of-variables formula, the identity $T^{-1}=T$, and Eq.~\eqref{eq:app_density_swap} give its unnormalized output density:
\begin{align*}
    r(T^{-1}(y))|\det DT^{-1}(y)|
    &=r(T(y))\\
    &=[q(T(y))-p(T(y))]_+\\
    &=[p(y)-q(y)]_+.
\end{align*}
Summing the accepted and corrected branches yields
\begin{equation}
    \min\{q(y),p(y)\}+[p(y)-q(y)]_+=p(y).
    \label{eq:app_mass_balance}
\end{equation}
Hence the final output has density $p$.
Furthermore, the rejection probability is $1-\int\min(p,q)$, which equals both $\int(q-p)_+$ and $\int(p-q)_+$ because $p$ and $q$ each integrate to one.
Conditioned on rejection, the corrected density is therefore
\begin{equation*}
    \frac{[p(y)-q(y)]_+}{\int[p(u)-q(u)]_+\,\mathrm du},
\end{equation*}
as required by Eq.~\eqref{eq:speculative_correction}.
\end{proof}

At a rejected step $j$, the direction $e$ is precisely $d_j$ in Eq.~\eqref{eq:coupling_reflection}.
The proof uses the original rejected candidate's noise: reflecting an independently redrawn Gaussian would instead produce the full target density on the rejection branch, rather than the required normalized residual.

\paragraph{Maximality.}
For any coupling $(\widetilde Y,Y)$ with marginal densities $q$ and $p$, define the diagonal subprobability measure $\nu(E)=\mathbb P(\widetilde Y=Y\in E)$ for measurable sets $E$.
It is dominated by both marginal measures, so it has a density bounded by both $q$ and $p$.
It follows that
\begin{equation*}
    \mathbb P(\widetilde Y=Y)\leq\int\min\{p(y),q(y)\}\,\mathrm dy.
\end{equation*}
The accepted branch of Lemma~\ref{lem:reflection} leaves the candidate unchanged with exactly this probability.
Together with the upper bound, this establishes maximality of the coupling.

\subsection{From a single corrected step to the complete path}
\label{app:path_proof}
\label{app:anchor_validity}

\begin{sagcproposition}[Exactness of the discrete speculative chain]
\label{prop:coupling_exact}
Under the model and arithmetic assumptions of Sec.~\ref{app:conditional_densities}, SAG-Coupling has the same joint law of $x_{1:K}$ as serial sampling from $P_k$, conditional on any fixed initial state $x_0$.
This holds for a prescribed positive noise schedule and history-dependent references, provided that verification uses the actual proposal density and each new proposal and acceptance test uses fresh independent randomness.
Consequently, for any common initial distribution $\mu_0$, both samplers also have the same joint law of $x_{0:K}$.
\end{sagcproposition}

\begin{proof}[Proof of Proposition~\ref{prop:coupling_exact}]
We first verify that batched computation and reference reuse meet the conditions of the single-step lemma, then identify the joint law.

\emph{Logical reveal of a verification window.}
At a window beginning at step $n$, the reference $\bar A$ is fixed before the window's random draws.
The recursion in Eq.~\eqref{eq:coupling_draft} makes each draft depend only on its predecessor, this reference, and its own Gaussian draw.
Consequently, the drafts can be generated with an independent sequence of Gaussian variables and tested with an independent sequence of uniform variables.
The statewise model assumption ensures that a target prediction at one draft does not depend on the other rows of the verification batch.
Precomputing later drafts and predictions therefore leaves every earlier candidate and acceptance decision unchanged.

Reveal the candidates and their acceptance variables in logical order.
At each tested transition $k$, all earlier candidates in the current window have been accepted, so $\widetilde x_k=x_k$.
The precomputed target mean is thus evaluated at the correct confirmed predecessor.
At the first rejection $j$, reflection uses this same predecessor and the noise $z_j$ of the rejected candidate, committing the corrected state $x_{j+1}$.
Later drafts are discarded, and the next window uses new independent random variables.
Unconsumed future draws are excluded from $\mathcal H_k$ and cannot be used to choose or revise the current draft mean.
This gives the same output path as batch computation while preserving the independence needed by Lemma~\ref{lem:reflection}.

\emph{Measurability of reused references.}
Initially, $\bar A=\widehat A_0(x_0)$ is a deterministic function of the initial state.
After full acceptance, the endpoint satisfies $x_{n+L}=\widetilde x_{n+L}$, and its cached prediction is a deterministic function of this confirmed state.
If $n+L<K$, its clean action estimate can therefore become the next reference without using any new-window randomness.

After rejection at step $j$, let $m=j+1<K$ and write $y=\widetilde x_m$ for the rejected candidate.
The new confirmed state is the reflected $x_m$, whereas the reference specified in the main text is
\begin{equation}
    \bar A=\widehat A_m(y)=y-\tau_m v_m(y).
    \label{eq:app_reused_reference}
\end{equation}
The candidate $y$ has already been revealed, and $v_m(y)$ is a deterministic function of $y$.
Thus, this reference is measurable with respect to the history at the start of the next window, even though it may differ from $\widehat A_m(x_m)$.
Fresh noise makes the new proposal conditionally Gaussian with its actual recorded mean.
Its target mean is computed from $v_m(x_m)$ at the corrected state; the cached $v_m(y)$ supplies only the reference and cannot replace that target evaluation.
Including the nonterminal draft endpoint in the verification batch makes Eq.~\eqref{eq:app_reused_reference} available even when the last candidate of a window is rejected.
Predictions from discarded later candidates are not used to choose this reference.
If $m=K$, sampling has ended and no new reference is required.

\emph{Conditional transition kernel.}
The preceding arguments show that, given $\mathcal H_k$, the current means are determined before the fresh Gaussian and uniform variables are revealed.
The two conditional densities are therefore exactly those in Eq.~\eqref{eq:app_conditional_densities}.
Applying Lemma~\ref{lem:reflection} conditionally gives, for every bounded measurable function $f$,
\begin{equation}
    \mathbb E[f(x_{k+1})\mid\mathcal H_k]
    =\int f(y)\,P_k(\mathrm dy\mid x_k).
    \label{eq:app_conditional_kernel}
\end{equation}
The right-hand side depends on the internal history only through $x_k$.
Because the confirmed path $x_{0:k}$ is contained in $\mathcal H_k$, the tower property gives
\begin{equation*}
    \begin{aligned}
        \mathbb E[f(x_{k+1})\mid x_{0:k}]
        &=\mathbb E\!\left[
            \mathbb E[f(x_{k+1})\mid\mathcal H_k]
            \mid x_{0:k}\right]\\
        &=\int f(y)\,P_k(\mathrm dy\mid x_k).
    \end{aligned}
\end{equation*}

\emph{Induction over the confirmed path.}
For any bounded measurable $g$ of the prefix $x_{0:k}$, the preceding identity implies
\begin{equation*}
    \begin{aligned}
        &\mathbb E[g(x_{0:k})f(x_{k+1})]\\
        &\quad=\mathbb E\!\left[g(x_{0:k})
                \int f(y)\,P_k(\mathrm dy\mid x_k)\right].
    \end{aligned}
\end{equation*}
If the prefix has the serial target law, the right-hand side is precisely its extension by one transition $P_k$.
Products of such test functions determine the joint distribution, so induction from the initial law $\mu_0$ yields
\begin{equation}
    \mathcal L(x_{0:K})
    =\mu_0(\mathrm dx_0)
        \prod_{k=0}^{K-1}P_k(\mathrm dx_{k+1}\mid x_k).
    \label{eq:app_joint_path_law}
\end{equation}
Each window commits at least one transition, including the corrected transition on rejection, so the algorithm reaches all $K$ steps.
Taking $\mu_0$ to be the point mass at a prescribed $x_0$ proves the conditional statement; integrating over any initial distribution proves the corresponding unconditional statement.
\end{proof}

The proposition concerns the discrete Gaussian chain defined by the learned velocity, time grid, and prescribed noise schedule.
It requires neither an exact score nor an accurate reference, and does not equate this chain with the continuous-time SDE or the original deterministic Euler sampler.
Its exact-arithmetic, statewise-model assumptions do not cover discrepancies caused by finite precision or batch-dependent predictions.

\subsection{Acceptance probability and the clean action reference}
\label{app:acceptance}

We next quantify how the clean action reference affects the probability of accepting a draft.
Conditioned on $\mathcal H_k$, the standardized mean distance $D=\|\Delta_k\|_2$ is fixed.
For $D>0$, set $e=\Delta_k/D$ and $S=e^\top z_k$.
Then $S\sim\mathcal N(0,1)$ and Eq.~\eqref{eq:app_log_ratio_derivation} becomes $\ell_k=-DS-D^2/2$.
The candidate is accepted with probability one when $S\leq-D/2$ and with probability $e^{-DS-D^2/2}$ otherwise.
Writing $\phi$ and $\Phi$ for the standard normal density and cumulative distribution function, completing the square gives
\begin{equation*}
    \phi(s)e^{-Ds-D^2/2}=\phi(s+D).
\end{equation*}
Integrating the two regions and substituting $u=s+D$ therefore yields
\begin{equation}
    \begin{aligned}
        \mathbb P(\mathrm{accept}\mid\mathcal H_k)
        &={}\int_{-\infty}^{-D/2}\phi(s)\,\mathrm ds\\
        &\quad+\int_{-D/2}^{\infty}
               \phi(s)e^{-Ds-D^2/2}\,\mathrm ds\\
        &=\Phi(-D/2)+\int_{D/2}^{\infty}\phi(u)\,\mathrm du\\
        &=2\Phi(-D/2).
    \end{aligned}
    \label{eq:app_acceptance_probability}
\end{equation}
For $D=0$, the means coincide and the same expression equals one.

To relate this probability to clean action drafting, consider the common predecessor $\widetilde x_k=x_k$ of a tested transition.
Using the estimate $\widehat A_k(x)=x-\tau_kv_k(x)$ and the reconstructed draft velocity from the main text gives
\begin{equation}
    \begin{aligned}
        \widetilde v_k-v_k(\widetilde x_k)
        &=\frac{\widetilde x_k-\bar A}{\tau_k}
             -v_k(\widetilde x_k)\\
        &=\frac{\widehat A_k(\widetilde x_k)-\bar A}{\tau_k}.
    \end{aligned}
    \label{eq:app_reference_velocity_difference}
\end{equation}
With the step coefficients defined in Appendix~\ref{app:discretization}, subtracting the target mean from the draft mean cancels their common state term, giving
\begin{equation*}
    \begin{aligned}
        m_k^Q-m_k^P
        &=-h_ka_k[\widetilde v_k-v_k(\widetilde x_k)]\\
        &=-\frac{h_ka_k}{\tau_k}
                [\widehat A_k(\widetilde x_k)-\bar A].
    \end{aligned}
\end{equation*}
Dividing by $\sigma_k$ and taking norms, with $h_k,a_k,\tau_k,\sigma_k>0$, gives
\begin{equation}
    \|\Delta_k\|_2
    =\frac{h_ka_k}{\tau_k\sigma_k}
        \|\widehat A_k(\widetilde x_k)-\bar A\|_2.
    \label{eq:app_reference_acceptance}
\end{equation}
At fixed step coefficients, a smaller discrepancy between the current clean action estimate and the reference gives a higher conditional acceptance probability.
Reference accuracy thus affects acceptance, while the distribution-preservation proof remains valid for any reference satisfying the stated history conditions.
Equation~\eqref{eq:app_acceptance_probability} describes a single transition conditioned on its history; it does not assert independence of acceptance events within a window.

\raggedbottom
\widowpenalty=10000
\clubpenalty=10000
\setlength{\intextsep}{6pt plus 1pt minus 1pt}
\setlength{\textfloatsep}{12pt plus 2pt minus 2pt}
\setlength{\dbltextfloatsep}{12pt plus 2pt minus 2pt}
\makeatletter
\setlength{\@fptop}{0pt}
\setlength{\@fpsep}{12pt plus 2pt minus 2pt}
\setlength{\@fpbot}{0pt plus 1fil}
\setlength{\@dblfptop}{0pt}
\setlength{\@dblfpsep}{12pt plus 2pt minus 2pt}
\setlength{\@dblfpbot}{0pt plus 1fil}
\makeatother
\section{Additional Simulation Results}
\label{app:simulation_details}

\subsection{Implementation and evaluation details}
\label{app:evaluation_details}

\paragraph{Models and inherited settings.}
We use the official Fast-WAM~\cite{yuanFastWAMWorldAction2026} and Faster-WAM~\cite{zhaoFasterWAMEfficientInferenceTime2026} checkpoints released by their authors.
For $\pi_{0.5}$~\cite{intelligence2025pi05visionlanguageactionmodelopenworld}, we use the official LIBERO checkpoint released by Physical Intelligence.
As Physical Intelligence has not publicly released a $\pi_{0.5}$ checkpoint for RoboTwin, we use the checkpoint released by the Motus authors~\cite{biMotusUnifiedLatent2025}.
This checkpoint was fine-tuned jointly on clean and randomized RoboTwin demonstrations, matching the task coverage and data composition used to train Fast-WAM and Faster-WAM.
We evaluate all three models on LIBERO-Plus using the same checkpoints as on LIBERO.
We retain all settings from each model's original release for the baseline.
When applying SAG, all other settings remain unchanged apart from SAG-specific sampling settings.
Table~\ref{tab:app_action_configuration} summarizes the action configurations.

\begin{table}[!htbp]
    \centering
    \caption{Inherited action configurations: chunk length $H$, sampler action dimension $D$, and executed length $E$. Quantile normalization uses the 1st and 99th percentiles.}
    \label{tab:app_action_configuration}
    \small
    \setlength{\tabcolsep}{3pt}
    \begin{tabular*}{\columnwidth}{@{\extracolsep{\fill}}lrrl@{}}
        \toprule
        Model & $H\!\times\!D$ & $E$ & Normalization \\
        \midrule
        \multicolumn{4}{@{}l}{\textbf{LIBERO / LIBERO-Plus}} \\
        $\pi_{0.5}$ & $10\!\times\!32$ & 5 & Quantile \\
        Fast-WAM & $32\!\times\!7$ & 10 & Min--max \\
        Faster-WAM & $32\!\times\!7$ & 10 & Min--max \\
        \midrule
        \multicolumn{4}{@{}l}{\textbf{RoboTwin}} \\
        $\pi_{0.5}$ & $32\!\times\!32$ & 32 & Quantile \\
        Fast-WAM & $32\!\times\!14$ & 24 & Z-score \\
        Faster-WAM & $32\!\times\!14$ & 28 & Z-score \\
        \bottomrule
    \end{tabular*}
\end{table}

\begin{table*}[t]
    \centering
    \caption{LIBERO inference latency (ms) by task suite. Avg. denotes the arithmetic mean across the four suites. Sampling parameters follow Table~\ref{tab:libero_results}.}
    \label{tab:app_libero_latency}
    \small
    \setlength{\tabcolsep}{6bp}
    \renewcommand{\arraystretch}{1.03}
    \begin{tabularx}{\textwidth}{>{\raggedright\arraybackslash}p{100bp}>{\raggedright\arraybackslash}p{45bp}*{5}{>{\raggedleft\arraybackslash}X}}
        \toprule
        Method & Stage & Spatial & Object & Goal & Long & Avg. \\
        \midrule
        \multicolumn{7}{c}{\textit{$\pi_{0.5}$}} \\
        \midrule
        Baseline & Action & 150.82 & 145.44 & 131.65 & 132.25 & 140.04 \\
         & Overall & 188.04 & 181.39 & 165.36 & 165.92 & 175.18 \\
        \rowcolor{black!4}
        \hspace*{3bp}+ SAG-Residual & Action & 69.53 & 65.06 & 59.27 & 59.49 & 63.34 \\
        \rowcolor{black!4}
         & Overall & 107.96 & 101.18 & 93.02 & 93.29 & 98.86 \\
        \rowcolor{black!4}
        \hspace*{3bp}+ SAG-Coupling & Action & 105.91 & 86.84 & 79.87 & 104.96 & 94.40 \\
        \rowcolor{black!4}
         & Overall & 142.82 & 123.41 & 114.75 & 137.91 & 129.72 \\
        \midrule
        \multicolumn{7}{c}{\textit{Fast-WAM}} \\
        \midrule
        Baseline & Action & 280.30 & 282.13 & 283.09 & 286.01 & 282.88 \\
         & Overall & 314.36 & 316.51 & 317.13 & 321.42 & 317.36 \\
        \rowcolor{black!4}
        \hspace*{3bp}+ SAG-Residual & Action & 125.30 & 114.81 & 134.84 & 133.72 & 127.17 \\
        \rowcolor{black!4}
         & Overall & 162.10 & 149.90 & 174.80 & 173.70 & 165.13 \\
        \rowcolor{black!4}
        \hspace*{3bp}+ SAG-Coupling & Action & 132.60 & 129.30 & 142.30 & 155.20 & 139.85 \\
        \rowcolor{black!4}
         & Overall & 166.70 & 165.90 & 177.50 & 192.30 & 175.60 \\
        \midrule
        \multicolumn{7}{c}{\textit{Faster-WAM}} \\
        \midrule
        Baseline & Action & 205.33 & 202.30 & 208.16 & 203.74 & 204.88 \\
         & Overall & 250.89 & 247.44 & 257.41 & 250.03 & 251.44 \\
        \rowcolor{black!4}
        \hspace*{3bp}+ SAG-Residual & Action & 135.32 & 103.42 & 80.41 & 86.35 & 101.38 \\
        \rowcolor{black!4}
         & Overall & 185.46 & 149.37 & 125.09 & 132.52 & 148.11 \\
        \rowcolor{black!4}
        \hspace*{3bp}+ SAG-Coupling & Action & 120.81 & 107.42 & 109.26 & 96.76 & 108.56 \\
        \rowcolor{black!4}
         & Overall & 170.29 & 153.71 & 160.49 & 144.41 & 157.23 \\
        \bottomrule
    \end{tabularx}
\end{table*}

\begin{table*}[!t]
    \centering
    \caption{Success rates and inference latencies of the Euler (ODE) baseline and serial SDE sampler for Faster-WAM on LIBERO.}
    \label{tab:app_libero_ode_sde}
    \small
    \setlength{\tabcolsep}{4pt}
    \begin{tabular*}{\textwidth}{@{\extracolsep{\fill}}lrrrrrrr@{}}
        \toprule
        & \multicolumn{5}{c}{Success rate (\%) $\uparrow$} & \multicolumn{2}{c}{Latency (ms) $\downarrow$} \\
        \cmidrule(lr){2-6}\cmidrule(l){7-8}
        Sampler & Spatial & Object & Goal & Long & Avg. & Action & Overall \\
        \midrule
        Serial Euler (ODE) & 99.4 & 99.6 & 98.2 & 98.2 & 98.85 & 204.88 & 251.44 \\
        Serial SDE & 98.4 & 99.8 & 98.4 & 98.0 & 98.65 & 206.19 & 253.01 \\
        \bottomrule
    \end{tabular*}
\end{table*}

\paragraph{Sampler implementation.}
We use each model's original time grid for both the baseline and SAG: a shifted grid with a shift parameter of 5 for the WAMs and a uniform grid for $\pi_{0.5}$,
\begin{equation}
    \begin{gathered}
    \tau_k =
    \begin{cases}
        \displaystyle\frac{5(1-k/K)}{1+4(1-k/K)}, & \text{Fast-WAM / Faster-WAM},\\[4pt]
        1-k/K, & \pi_{0.5},
    \end{cases}\\
    k=0,\ldots,K.
    \end{gathered}
    \label{eq:app_inherited_time_grid}
\end{equation}
For SAG-Residual, we compute the $L_2$ norm in Eq.~\eqref{eq:residual_test} after flattening the $H\times D$ difference tensor in normalized action space, without dimension normalization.
The WAMs use their native action dimensions without padding.
For $\pi_{0.5}$, all 32 action dimensions are included without masking.
Padded coordinates can contribute to the residual at intermediate denoising steps.
All models use BF16 model computation.
For the WAMs, sampling states use BF16 for the baseline and SAG-Residual, and FP32 for SAG-Coupling and the serial SDE control.
For $\pi_{0.5}$, the baseline and both SAG variants maintain sampling states in FP32.

\paragraph{Evaluation protocol.}
We follow the evaluation settings in each model's official release.
On LIBERO, we use the official initial states.
For the WAMs, episode limits are 400 steps (700 for Long).
For $\pi_{0.5}$, the limits are 220, 280, 300, and 520 steps for Spatial, Object, Goal, and Long, respectively.
The WAMs also retain 30 initial steps for the environment to settle.
Following Faster-WAM~\cite{zhaoFasterWAMEfficientInferenceTime2026}, we evaluate on a fixed, randomly sampled 15\% subset of LIBERO-Plus (1,506 task instances), with one rollout per instance.
On RoboTwin, we retain the original evaluation setup with unseen language instructions and report the arithmetic mean of the clean and randomized success rates.

\paragraph{Timing details.}
We measure latency on a single GPU using one model instance in eager mode, with requests processed sequentially.
Each request contains one observation (batch size 1), while SAG batches candidate states internally for verification.
Text embeddings are cached before timing, and measurements exclude image preprocessing, text encoding, action denormalization, and gripper postprocessing.

\paragraph{Profiling inputs.}
We profile latency separately from success rate evaluation, using saved observations shared across samplers within each model.
On LIBERO, all three models use the same four initial observations, one per suite from the first task and its first official initial state.
The baseline and SAG-Residual use one random stream per observation, while SAG-Coupling uses eight.
We average repeated measurements within each stream, then across streams, and finally equally across the four suites.
LIBERO-Plus latency is measured separately using the same protocol.

On RoboTwin, all three models use nine observations from selected tasks, six clean and three randomized, collected at the first policy request of the first episode.
We average repeated measurements within each observation and then equally across the nine observations.

\subsection{Additional LIBERO results}
\label{app:libero_results}
\paragraph{Latency by task suite.}
To complement Table~\ref{tab:libero_results}, Table~\ref{tab:app_libero_latency} reports action generation and overall inference latency for each LIBERO task suite.

\paragraph{ODE-to-SDE conversion.}
SAG-Coupling first converts the action ODE into an SDE and then applies speculative sampling to its discretized transitions.
To assess the impact of this reformulation on task performance and latency, we disable speculation in SAG-Coupling and retain all other settings.
Table~\ref{tab:app_libero_ode_sde} compares this serial SDE sampler with the original Euler (ODE) baseline for Faster-WAM on LIBERO.
The two samplers have comparable average success rates and similar action generation and overall inference latencies.

\paragraph{Action deviations.}
To further assess the impact of SAG on action quality, we compare the action chunks generated by both variants with the baseline outputs.
\begin{table}[!t]
    \centering
    \caption{Action deviations for Faster-WAM on LIBERO. Both panels report mean $L_2$ distances between flattened, normalized action chunks, with Avg. denoting the mean across the four suites. (a) Comparison of both SAG variants with the baseline during evaluation. (b) Comparison of the baseline and both SAG variants with the ground truth on the dataset.}
    \label{tab:app_libero_action_deviation}
    \small
    \setlength{\tabcolsep}{1.8bp}
    \renewcommand{\arraystretch}{1.03}
    \begin{tabularx}{\columnwidth}{>{\raggedright\arraybackslash}p{65bp}*{5}{>{\raggedleft\arraybackslash}X}}
        \toprule
        Method & Spatial & Object & Goal & Long & Avg. \\
        \midrule
        \multicolumn{6}{c}{\textit{(a) Deviation from the baseline}} \\
        \midrule
        \rowcolor{black!4}
        SAG-Residual & 0.0534 & 0.0308 & 0.0298 & 0.0269 & 0.0352 \\
        \rowcolor{black!4}
        SAG-Coupling & 0.5500 & 0.5330 & 0.5381 & 0.5346 & 0.5389 \\
        \midrule
        \multicolumn{6}{c}{\textit{(b) Deviation from dataset ground truth}} \\
        \midrule
        Baseline & 1.8545 & 1.5305 & 1.6350 & 1.9237 & 1.7359 \\
        \rowcolor{black!4}
        SAG-Residual & 1.8505 & 1.5270 & 1.6364 & 1.9254 & 1.7348 \\
        \rowcolor{black!4}
        SAG-Coupling & 1.9134 & 1.6434 & 1.7204 & 2.0075 & 1.8212 \\
        \bottomrule
    \end{tabularx}
\end{table}
We evaluate Faster-WAM on 16 observations collected during evaluation near the initial state of the first task in each LIBERO suite, using the same observations and initial action noise for all methods.
As shown in Table~\ref{tab:app_libero_action_deviation}(a), both SAG variants exhibit only minor deviations from the baseline action outputs.
In a separate experiment on the LIBERO dataset, we compare the generated action chunks with the corresponding ground truth actions.
Table~\ref{tab:app_libero_action_deviation}(b) shows that both SAG variants introduce only minor changes in action prediction error.

\subsection{RoboTwin task-level results}
\label{app:robotwin_tasks}
The main paper reports average success rates on RoboTwin~2.0 under the clean and randomized evaluation settings.
Here, we provide the corresponding per-task success rates for all three baselines and their SAG variants.
Detailed results for $\pi_{0.5}$, Fast-WAM, and Faster-WAM are presented in Tables~\ref{tab:app_pi05_robotwin_tasks}--\ref{tab:app_fasterwam_robotwin_tasks}, respectively.

\begin{table}[!tp]
    \centering
    \caption{Per-task success rates (\%) for $\pi_{0.5}$ and its SAG variants on RoboTwin~2.0 under the clean and randomized settings. Avg. denotes the mean across all 50 tasks.}
    \label{tab:app_pi05_robotwin_tasks}
    \fontsize{7.8}{9.2}\selectfont
    \renewcommand{\arraystretch}{1.02}
    \setlength{\tabcolsep}{0pt}
    \begin{tabular}{@{}>{\raggedright\arraybackslash}p{\dimexpr\columnwidth-153bp\relax}
        @{\hspace{7bp}}>{\raggedleft\arraybackslash}p{21bp}
        @{\hspace{2bp}}>{\raggedleft\arraybackslash}p{21bp}
        @{\hspace{7bp}}>{\raggedleft\arraybackslash}p{21bp}
        @{\hspace{2bp}}>{\raggedleft\arraybackslash}p{21bp}
        @{\hspace{7bp}}>{\raggedleft\arraybackslash}p{21bp}
        @{\hspace{2bp}}>{\raggedleft\arraybackslash}p{21bp}@{}}
\toprule
        & \multicolumn{2}{c}{$\pi_{0.5}$} & \multicolumn{2}{c}{\parbox[c]{44bp}{\centering + SAG-Residual}} & \multicolumn{2}{c}{\parbox[c]{44bp}{\centering + SAG-Coupling}} \\
        \cmidrule(l{1bp}r{1bp}){2-3}\cmidrule(l{1bp}r{1bp}){4-5}\cmidrule(l{1bp}r{1bp}){6-7}
        Task & \makebox[21bp][c]{Clean} & \makebox[21bp][c]{Rand.} & \makebox[21bp][c]{Clean} & \makebox[21bp][c]{Rand.} & \makebox[21bp][c]{Clean} & \makebox[21bp][c]{Rand.} \\
        \midrule
        Adjust Bottle & 95 & 95 & 98 & 96 & 95 & 88 \\
        Beat Block Hammer & 56 & 65 & 91 & 91 & 72 & 68 \\
        Blocks Ranking RGB & 43 & 44 & 60 & 69 & 42 & 43 \\
        Blocks Ranking Size & 22 & 23 & 23 & 24 & 18 & 22 \\
        Click Alarmclock & 68 & 63 & 71 & 69 & 71 & 59 \\
        Click Bell & 55 & 53 & 44 & 65 & 61 & 59 \\
        Dump Bin Bigbin & 70 & 61 & 74 & 78 & 59 & 56 \\
        Grab Roller & 88 & 89 & 97 & 98 & 82 & 87 \\
        Handover Block & 11 & 8 & 15 & 21 & 8 & 12 \\
        Handover Mic & 62 & 51 & 61 & 54 & 63 & 49 \\
        Hanging Mug & 5 & 4 & 9 & 5 & 2 & 5 \\
        Lift Pot & 29 & 23 & 33 & 27 & 28 & 20 \\
        Move Can Pot & 29 & 33 & 34 & 54 & 31 & 39 \\
        Move Pillbottle Pad & 32 & 40 & 36 & 54 & 30 & 45 \\
        Move Playingcard Away & 78 & 87 & 80 & 84 & 76 & 74 \\
        Move Stapler Pad & 14 & 19 & 23 & 27 & 10 & 13 \\
        Open Laptop & 38 & 39 & 43 & 44 & 38 & 31 \\
        Open Microwave & 60 & 46 & 59 & 51 & 45 & 48 \\
        Pick Diverse Bottles & 35 & 31 & 56 & 56 & 42 & 33 \\
        Pick Dual Bottles & 38 & 40 & 53 & 56 & 31 & 35 \\
        Place A2B Left & 56 & 61 & 58 & 68 & 58 & 57 \\
        Place A2B Right & 63 & 68 & 64 & 62 & 51 & 67 \\
        Place Bread Basket & 38 & 62 & 44 & 66 & 34 & 57 \\
        Place Bread Skillet & 29 & 27 & 32 & 38 & 27 & 31 \\
        Place Burger Fries & 62 & 55 & 60 & 75 & 57 & 71 \\
        Place Can Basket & 20 & 32 & 45 & 33 & 35 & 26 \\
        Place Cans Plasticbox & 20 & 32 & 22 & 37 & 28 & 36 \\
        Place Container Plate & 71 & 83 & 80 & 91 & 73 & 83 \\
        Place Dual Shoes & 21 & 32 & 35 & 41 & 25 & 29 \\
        Place Empty Cup & 54 & 67 & 26 & 35 & 53 & 62 \\
        Place Fan & 37 & 49 & 38 & 56 & 29 & 47 \\
        Place Mouse Pad & 24 & 24 & 18 & 27 & 13 & 28 \\
        Place Object Basket & 39 & 50 & 56 & 51 & 38 & 49 \\
        Place Object Scale & 40 & 54 & 48 & 70 & 50 & 49 \\
        Place Object Stand & 63 & 77 & 82 & 82 & 65 & 69 \\
        Place Phone Stand & 46 & 49 & 46 & 58 & 48 & 55 \\
        Place Shoe & 55 & 59 & 62 & 70 & 50 & 49 \\
        Press Stapler & 58 & 49 & 65 & 70 & 58 & 52 \\
        Put Bottles Dustbin & 24 & 27 & 32 & 37 & 22 & 25 \\
        Put Object Cabinet & 20 & 17 & 34 & 32 & 19 & 19 \\
        Rotate QR Code & 37 & 35 & 46 & 47 & 41 & 39 \\
        Scan Object & 17 & 23 & 28 & 30 & 16 & 17 \\
        Shake Bottle & 96 & 96 & 97 & 98 & 95 & 95 \\
        Shake Bottle Horizontally & 95 & 97 & 98 & 100 & 93 & 93 \\
        Stack Blocks Three & 19 & 20 & 27 & 42 & 18 & 28 \\
        Stack Blocks Two & 52 & 47 & 62 & 71 & 49 & 53 \\
        Stack Bowls Three & 34 & 38 & 53 & 46 & 42 & 34 \\
        Stack Bowls Two & 74 & 82 & 77 & 82 & 74 & 79 \\
        Stamp Seal & 42 & 37 & 37 & 37 & 33 & 30 \\
        Turn Switch & 23 & 30 & 27 & 36 & 17 & 31 \\
        \midrule
        Avg. & 45.14 & 47.86 & 51.18 & 56.22 & 44.30 & 46.92 \\
        \bottomrule
    \end{tabular}
\end{table}

\begin{table}[!tp]
    \centering
    \caption{Per-task success rates (\%) for Fast-WAM and its SAG variants on RoboTwin~2.0 under the clean and randomized settings. Avg. denotes the mean across all 50 tasks.}
    \label{tab:app_robotwin_tasks}
    \fontsize{7.8}{9.2}\selectfont
    \renewcommand{\arraystretch}{1.02}
    \setlength{\tabcolsep}{0pt}
    \begin{tabular}{@{}>{\raggedright\arraybackslash}p{\dimexpr\columnwidth-153bp\relax}
        @{\hspace{7bp}}>{\raggedleft\arraybackslash}p{21bp}
        @{\hspace{2bp}}>{\raggedleft\arraybackslash}p{21bp}
        @{\hspace{7bp}}>{\raggedleft\arraybackslash}p{21bp}
        @{\hspace{2bp}}>{\raggedleft\arraybackslash}p{21bp}
        @{\hspace{7bp}}>{\raggedleft\arraybackslash}p{21bp}
        @{\hspace{2bp}}>{\raggedleft\arraybackslash}p{21bp}@{}}
\toprule
        & \multicolumn{2}{c}{Fast-WAM} & \multicolumn{2}{c}{\parbox[c]{44bp}{\centering + SAG-Residual}} & \multicolumn{2}{c}{\parbox[c]{44bp}{\centering + SAG-Coupling}} \\
        \cmidrule(l{1bp}r{1bp}){2-3}\cmidrule(l{1bp}r{1bp}){4-5}\cmidrule(l{1bp}r{1bp}){6-7}
        Task & \makebox[21bp][c]{Clean} & \makebox[21bp][c]{Rand.} & \makebox[21bp][c]{Clean} & \makebox[21bp][c]{Rand.} & \makebox[21bp][c]{Clean} & \makebox[21bp][c]{Rand.} \\
        \midrule
        Adjust Bottle & 100 & 100 & 100 & 100 & 100 & 100 \\
        Beat Block Hammer & 100 & 98 & 98 & 98 & 99 & 99 \\
        Blocks Ranking RGB & 100 & 99 & 100 & 100 & 100 & 99 \\
        Blocks Ranking Size & 92 & 94 & 95 & 96 & 94 & 91 \\
        Click Alarmclock & 100 & 100 & 100 & 100 & 100 & 100 \\
        Click Bell & 100 & 100 & 100 & 100 & 100 & 100 \\
        Dump Bin Bigbin & 97 & 98 & 96 & 98 & 97 & 95 \\
        Grab Roller & 100 & 100 & 100 & 100 & 100 & 100 \\
        Handover Block & 91 & 83 & 93 & 82 & 94 & 78 \\
        Handover Mic & 99 & 100 & 100 & 100 & 98 & 98 \\
        Hanging Mug & 59 & 58 & 64 & 58 & 59 & 65 \\
        Lift Pot & 100 & 100 & 99 & 100 & 99 & 100 \\
        Move Can Pot & 92 & 87 & 90 & 91 & 87 & 90 \\
        Move Pillbottle Pad & 98 & 99 & 99 & 99 & 99 & 99 \\
        Move Playingcard Away & 100 & 100 & 100 & 100 & 100 & 100 \\
        Move Stapler Pad & 77 & 75 & 81 & 72 & 80 & 64 \\
        Open Laptop & 53 & 58 & 53 & 58 & 53 & 58 \\
        Open Microwave & 84 & 53 & 83 & 47 & 84 & 57 \\
        Pick Diverse Bottles & 84 & 85 & 84 & 84 & 83 & 84 \\
        Pick Dual Bottles & 100 & 97 & 100 & 96 & 99 & 97 \\
        Place A2B Left & 97 & 96 & 96 & 95 & 94 & 93 \\
        Place A2B Right & 98 & 94 & 97 & 94 & 98 & 93 \\
        Place Bread Basket & 90 & 94 & 92 & 95 & 87 & 92 \\
        Place Bread Skillet & 94 & 93 & 95 & 93 & 94 & 88 \\
        Place Burger Fries & 98 & 99 & 96 & 95 & 97 & 98 \\
        Place Can Basket & 68 & 60 & 68 & 64 & 62 & 55 \\
        Place Cans Plasticbox & 98 & 99 & 99 & 97 & 99 & 98 \\
        Place Container Plate & 100 & 99 & 97 & 97 & 98 & 95 \\
        Place Dual Shoes & 92 & 90 & 88 & 88 & 91 & 88 \\
        Place Empty Cup & 100 & 100 & 100 & 100 & 100 & 100 \\
        Place Fan & 95 & 94 & 95 & 93 & 93 & 91 \\
        Place Mouse Pad & 84 & 90 & 85 & 89 & 91 & 86 \\
        Place Object Basket & 82 & 85 & 85 & 90 & 81 & 91 \\
        Place Object Scale & 97 & 94 & 94 & 95 & 92 & 92 \\
        Place Object Stand & 89 & 87 & 88 & 91 & 88 & 88 \\
        Place Phone Stand & 98 & 99 & 99 & 98 & 98 & 98 \\
        Place Shoe & 97 & 98 & 95 & 99 & 95 & 99 \\
        Press Stapler & 91 & 97 & 92 & 97 & 93 & 97 \\
        Put Bottles Dustbin & 94 & 87 & 94 & 87 & 93 & 90 \\
        Put Object Cabinet & 46 & 49 & 46 & 49 & 47 & 46 \\
        Rotate QR Code & 92 & 88 & 93 & 91 & 96 & 86 \\
        Scan Object & 91 & 88 & 90 & 92 & 92 & 93 \\
        Shake Bottle & 100 & 100 & 100 & 100 & 100 & 100 \\
        Shake Bottle Horizontally & 100 & 100 & 100 & 100 & 100 & 100 \\
        Stack Blocks Three & 96 & 96 & 96 & 96 & 98 & 95 \\
        Stack Blocks Two & 100 & 100 & 100 & 100 & 99 & 99 \\
        Stack Bowls Three & 82 & 84 & 79 & 84 & 77 & 77 \\
        Stack Bowls Two & 93 & 97 & 93 & 96 & 95 & 95 \\
        Stamp Seal & 88 & 94 & 90 & 93 & 90 & 93 \\
        Turn Switch & 66 & 64 & 61 & 62 & 58 & 64 \\
        \midrule
        Avg. & 90.84 & 89.98 & 90.76 & 89.98 & 90.42 & 89.08 \\
        \bottomrule
    \end{tabular}
\end{table}

\begin{table}[!tp]
    \centering
    \caption{Per-task success rates (\%) for Faster-WAM and its SAG variants on RoboTwin~2.0 under the clean and randomized settings. Avg. denotes the mean across all 50 tasks.}
    \label{tab:app_fasterwam_robotwin_tasks}
    \fontsize{7.8}{9.2}\selectfont
    \renewcommand{\arraystretch}{1.02}
    \setlength{\tabcolsep}{0pt}
    \begin{tabular}{@{}>{\raggedright\arraybackslash}p{\dimexpr\columnwidth-153bp\relax}
        @{\hspace{7bp}}>{\raggedleft\arraybackslash}p{21bp}
        @{\hspace{2bp}}>{\raggedleft\arraybackslash}p{21bp}
        @{\hspace{7bp}}>{\raggedleft\arraybackslash}p{21bp}
        @{\hspace{2bp}}>{\raggedleft\arraybackslash}p{21bp}
        @{\hspace{7bp}}>{\raggedleft\arraybackslash}p{21bp}
        @{\hspace{2bp}}>{\raggedleft\arraybackslash}p{21bp}@{}}
\toprule
        & \multicolumn{2}{c}{Faster-WAM} & \multicolumn{2}{c}{\parbox[c]{44bp}{\centering + SAG-Residual}} & \multicolumn{2}{c}{\parbox[c]{44bp}{\centering + SAG-Coupling}} \\
        \cmidrule(l{1bp}r{1bp}){2-3}\cmidrule(l{1bp}r{1bp}){4-5}\cmidrule(l{1bp}r{1bp}){6-7}
        Task & \makebox[21bp][c]{Clean} & \makebox[21bp][c]{Rand.} & \makebox[21bp][c]{Clean} & \makebox[21bp][c]{Rand.} & \makebox[21bp][c]{Clean} & \makebox[21bp][c]{Rand.} \\
        \midrule
        Adjust Bottle & 100 & 99 & 100 & 99 & 100 & 99 \\
        Beat Block Hammer & 100 & 99 & 100 & 98 & 100 & 98 \\
        Blocks Ranking RGB & 99 & 100 & 100 & 99 & 99 & 100 \\
        Blocks Ranking Size & 79 & 92 & 78 & 92 & 77 & 91 \\
        Click Alarmclock & 100 & 100 & 99 & 100 & 100 & 100 \\
        Click Bell & 100 & 100 & 100 & 100 & 100 & 100 \\
        Dump Bin Bigbin & 99 & 96 & 98 & 98 & 98 & 96 \\
        Grab Roller & 100 & 100 & 100 & 100 & 100 & 100 \\
        Handover Block & 94 & 91 & 93 & 90 & 92 & 88 \\
        Handover Mic & 100 & 99 & 100 & 100 & 100 & 99 \\
        Hanging Mug & 51 & 50 & 57 & 50 & 52 & 49 \\
        Lift Pot & 100 & 100 & 100 & 100 & 100 & 100 \\
        Move Can Pot & 96 & 98 & 97 & 97 & 97 & 97 \\
        Move Pillbottle Pad & 96 & 99 & 98 & 99 & 99 & 99 \\
        Move Playingcard Away & 100 & 99 & 100 & 99 & 100 & 99 \\
        Move Stapler Pad & 80 & 79 & 79 & 72 & 79 & 76 \\
        Open Laptop & 53 & 55 & 53 & 55 & 53 & 55 \\
        Open Microwave & 78 & 74 & 79 & 78 & 77 & 62 \\
        Pick Diverse Bottles & 90 & 88 & 88 & 88 & 89 & 89 \\
        Pick Dual Bottles & 99 & 95 & 99 & 95 & 99 & 95 \\
        Place A2B Left & 95 & 97 & 96 & 95 & 94 & 94 \\
        Place A2B Right & 94 & 94 & 96 & 95 & 96 & 96 \\
        Place Bread Basket & 92 & 96 & 92 & 96 & 92 & 95 \\
        Place Bread Skillet & 92 & 87 & 97 & 91 & 95 & 94 \\
        Place Burger Fries & 100 & 98 & 99 & 99 & 98 & 98 \\
        Place Can Basket & 65 & 68 & 76 & 74 & 72 & 66 \\
        Place Cans Plasticbox & 99 & 94 & 99 & 96 & 97 & 97 \\
        Place Container Plate & 99 & 99 & 100 & 98 & 95 & 98 \\
        Place Dual Shoes & 89 & 90 & 88 & 88 & 90 & 84 \\
        Place Empty Cup & 100 & 100 & 100 & 100 & 100 & 100 \\
        Place Fan & 99 & 100 & 97 & 96 & 97 & 97 \\
        Place Mouse Pad & 86 & 95 & 86 & 93 & 87 & 92 \\
        Place Object Basket & 80 & 83 & 82 & 88 & 85 & 82 \\
        Place Object Scale & 96 & 98 & 96 & 99 & 96 & 98 \\
        Place Object Stand & 98 & 94 & 97 & 97 & 96 & 92 \\
        Place Phone Stand & 99 & 94 & 100 & 94 & 100 & 95 \\
        Place Shoe & 96 & 98 & 96 & 98 & 95 & 99 \\
        Press Stapler & 92 & 94 & 93 & 95 & 93 & 94 \\
        Put Bottles Dustbin & 96 & 87 & 95 & 89 & 93 & 91 \\
        Put Object Cabinet & 44 & 46 & 43 & 45 & 45 & 45 \\
        Rotate QR Code & 92 & 88 & 91 & 90 & 93 & 92 \\
        Scan Object & 88 & 90 & 88 & 92 & 86 & 91 \\
        Shake Bottle & 100 & 100 & 100 & 100 & 100 & 100 \\
        Shake Bottle Horizontally & 100 & 100 & 100 & 100 & 100 & 100 \\
        Stack Blocks Three & 98 & 98 & 97 & 97 & 97 & 97 \\
        Stack Blocks Two & 100 & 100 & 100 & 100 & 100 & 100 \\
        Stack Bowls Three & 78 & 79 & 87 & 88 & 83 & 78 \\
        Stack Bowls Two & 95 & 96 & 94 & 93 & 99 & 97 \\
        Stamp Seal & 91 & 90 & 92 & 91 & 87 & 89 \\
        Turn Switch & 74 & 78 & 73 & 75 & 74 & 77 \\
        \midrule
        Avg. & 90.82 & 90.88 & 91.36 & 91.22 & 90.92 & 90.40 \\
        \bottomrule
    \end{tabular}
\end{table}

\subsection{Component-wise inference latency}
\label{app:latency_breakdown}
To complement the latency results in the main paper, we report component-wise inference latency for all three baselines and their SAG variants.
Tables~\ref{tab:app_libero_components} and~\ref{tab:app_robotwin_components} present the results on LIBERO and RoboTwin~2.0, respectively, including VAE encoding, visual processing, and action generation alongside overall inference latency.

\begin{table}[!htbp]
    \centering
    \caption{Component-wise inference latency on LIBERO (ms). Settings follow Table~\ref{tab:libero_results}. Overall is timed from prepared inputs to the output action chunk. A dash indicates that the model has no VAE encoding stage.}
    \label{tab:app_libero_components}
    \small
    \setlength{\tabcolsep}{3pt}
    \renewcommand{\arraystretch}{1.05}
    \begin{tabular*}{\columnwidth}{@{\extracolsep{\fill}}lrrrr@{}}
        \toprule
        Method & VAE enc. & Visual & Action & Overall \\
        \midrule
        \multicolumn{5}{@{}l}{\textbf{$\pi_{0.5}$}} \\
        Baseline & --- & 34.99 & 140.04 & 175.18 \\
        + SAG-Residual & --- & 35.40 & 63.34 & 98.86 \\
        + SAG-Coupling & --- & 35.19 & 94.40 & 129.72 \\
        \midrule
        \multicolumn{5}{@{}l}{\textbf{Fast-WAM}} \\
        Baseline & 5.68 & 28.42 & 282.88 & 317.36 \\
        + SAG-Residual & 6.18 & 31.40 & 127.17 & 165.13 \\
        + SAG-Coupling & 5.78 & 29.68 & 139.85 & 175.60 \\
        \midrule
        \multicolumn{5}{@{}l}{\textbf{Faster-WAM}} \\
        Baseline & 6.48 & 32.29 & 204.88 & 251.44 \\
        + SAG-Residual & 6.21 & 32.80 & 101.38 & 148.11 \\
        + SAG-Coupling & 6.57 & 33.97 & 108.56 & 157.23 \\
        \bottomrule
    \end{tabular*}
\end{table}

\begin{table}[!htbp]
    \centering
    \caption{Component-wise inference latency on RoboTwin~2.0 (ms). Settings follow Table~\ref{tab:robotwin_results}. Overall is timed from prepared inputs to the output action chunk. A dash indicates that the model has no VAE encoding stage.}
    \label{tab:app_robotwin_components}
    \small
    \setlength{\tabcolsep}{3pt}
    \renewcommand{\arraystretch}{1.05}
    \begin{tabular*}{\columnwidth}{@{\extracolsep{\fill}}lrrrr@{}}
        \toprule
        Method & VAE enc. & Visual & Action & Overall \\
        \midrule
        \multicolumn{5}{@{}l}{\textbf{$\pi_{0.5}$}} \\
        Baseline & --- & 56.29 & 186.14 & 242.54 \\
        + SAG-Residual & --- & 56.96 & 65.16 & 122.25 \\
        + SAG-Coupling & --- & 56.42 & 133.68 & 190.23 \\
        \midrule
        \multicolumn{5}{@{}l}{\textbf{Fast-WAM}} \\
        Baseline & 8.90 & 43.45 & 428.95 & 481.69 \\
        + SAG-Residual & 8.93 & 43.64 & 132.53 & 185.47 \\
        + SAG-Coupling & 8.27 & 41.79 & 201.13 & 251.51 \\
        \midrule
        \multicolumn{5}{@{}l}{\textbf{Faster-WAM}} \\
        Baseline & 8.67 & 43.06 & 286.41 & 346.86 \\
        + SAG-Residual & 8.52 & 43.52 & 87.52 & 148.19 \\
        + SAG-Coupling & 8.74 & 42.54 & 120.37 & 180.37 \\
        \bottomrule
    \end{tabular*}
\end{table}

\subsection{Ablation Studies}
\label{app:robotwin_subset_studies}
\label{app:robotwin_parameter_results}

\paragraph{Task selection.}
We conduct ablation studies in the more challenging randomized setting of RoboTwin~2.0 to examine the effects of SAG hyperparameters and core components.
To this end, we select six challenging tasks with substantial variation in denoising velocity, while limiting ceiling effects from high baseline success rates.
For each of the 50 RoboTwin~2.0 tasks, we sample 20 episodes from the offline dataset and use the first observation of each episode.
We run the Faster-WAM baseline on these observations and record the denoising velocities $v_k^{(e)}$.
We score task $\mathcal T$ by the mean relative change between consecutive velocities,
\begingroup
\setlength{\abovedisplayskip}{6pt}
\setlength{\belowdisplayskip}{6pt}
\begin{equation}
    C(\mathcal T)=\operatorname{mean}_{e,k}
    \frac{\|v_{k+1}^{(e)}-v_k^{(e)}\|_2}
         {\|v_{k+1}^{(e)}\|_2},
    \label{eq:app_robotwin_screening_score}
\end{equation}
\endgroup
where the mean is taken over sampled episodes $e$ and adjacent integration steps $k$.
We then rank these tasks by decreasing score, exclude those with baseline success rates of $90\%$ or higher, and select the first six remaining tasks.
Table~\ref{tab:app_robotwin_selection} summarizes the resulting task selection.

\begingroup
\setlength{\intextsep}{4pt}
\begin{table}[!htbp]
\centering
\captionsetup{skip=4pt}
\caption{Task selection for the ablation studies. The top 11 ranked tasks are shown, with the six selected tasks in bold.}
\label{tab:app_robotwin_selection}
\small
\setlength{\tabcolsep}{3pt}
\renewcommand{\arraystretch}{1.04}
\begin{tabular*}{\linewidth}{@{\extracolsep{\fill}}rlrr@{}}
\toprule
Rank & Task & Score & SR (\%) \\
\midrule
1 & Pick dual bottles & 0.0103 & 97.0 \\
\textbf{2} & \textbf{Open microwave} & \textbf{0.0102} & \textbf{76.0} \\
\textbf{3} & \textbf{Pick diverse bottles} & \textbf{0.0099} & \textbf{89.0} \\
\textbf{4} & \textbf{Open laptop} & \textbf{0.0096} & \textbf{54.0} \\
5 & Grab roller & 0.0094 & 100.0 \\
6 & Adjust bottle & 0.0094 & 99.5 \\
\textbf{7} & \textbf{Turn switch} & \textbf{0.0094} & \textbf{76.0} \\
8 & Put bottles dustbin & 0.0094 & 91.5 \\
\textbf{9} & \textbf{Scan object} & \textbf{0.0093} & \textbf{89.0} \\
10 & Rotate qrcode & 0.0091 & 90.0 \\
\textbf{11} & \textbf{Hanging mug} & \textbf{0.0090} & \textbf{50.5} \\
\bottomrule
\end{tabular*}
\end{table}
\endgroup

\begin{table}[!htbp]
\centering
\captionsetup{skip=4pt}
\caption{Hyperparameter ablation results for Faster-WAM on the six selected tasks. Rand. denotes the randomized setting. Pooled is the mean of the clean and randomized success rates. Bold rows indicate the configurations used in the main experiments.}
\label{tab:app_robotwin_parameter_summary}
\fontsize{9}{10.4}\selectfont
\setlength{\tabcolsep}{2pt}
\renewcommand{\arraystretch}{1.00}
\begin{tabular*}{\linewidth}{@{\extracolsep{\fill}}rrrrrrr@{}}
\toprule
& & \multicolumn{3}{c}{Success rate (\%) $\uparrow$} & \multicolumn{2}{c}{Latency (ms) $\downarrow$} \\
\cmidrule(lr){3-5}\cmidrule(l){6-7}
$M$ & $\delta$ / $\lambda$ & Clean & Rand. & Pooled & Action & Overall \\
\midrule
\multicolumn{7}{l}{\textbf{SAG-Residual} ($\delta$)} \\
\midrule
2 & 0.05 & 73.17 & 74.17 & 73.67 & 181.49 & 244.38 \\
2 & 0.1 & 73.50 & 71.17 & 72.33 & 180.58 & 244.51 \\
2 & 0.2 & 73.33 & 69.33 & 71.33 & 176.49 & 241.36 \\
4 & 0.05 & 73.17 & 73.17 & 73.17 & 122.54 & 189.16 \\
4 & 0.1 & 73.00 & 70.83 & 71.92 & 118.67 & 182.78 \\
4 & 0.2 & 73.00 & 69.00 & 71.00 & 120.44 & 183.01 \\
8 & 0.05 & 72.83 & 72.83 & 72.83 & 125.92 & 190.71 \\
\textbf{8} & \textbf{0.1} & \textbf{73.00} & \textbf{72.83} & \textbf{72.92} & \textbf{87.17} & \textbf{147.85} \\
8 & 0.2 & 73.17 & 72.50 & 72.83 & 92.99 & 158.08 \\
10 & 0.05 & 72.83 & 72.33 & 72.58 & 125.95 & 191.32 \\
10 & 0.1 & 72.50 & 72.33 & 72.42 & 89.64 & 151.02 \\
10 & 0.2 & 72.00 & 74.17 & 73.08 & 93.05 & 157.85 \\
\midrule
\multicolumn{7}{l}{\textbf{SAG-Coupling} ($\lambda$)} \\
\midrule
2 & 0.05 & 72.67 & 70.83 & 71.75 & 218.67 & 281.26 \\
2 & 0.1 & 72.67 & 70.17 & 71.42 & 182.78 & 250.92 \\
2 & 0.2 & 73.50 & 71.83 & 72.67 & 186.69 & 253.52 \\
4 & 0.05 & 72.67 & 72.17 & 72.42 & 186.97 & 249.33 \\
\textbf{4} & \textbf{0.1} & \textbf{71.83} & \textbf{70.50} & \textbf{71.17} & \textbf{133.31} & \textbf{199.06} \\
4 & 0.2 & 72.33 & 69.83 & 71.08 & 123.62 & 184.92 \\
8 & 0.05 & 73.67 & 70.83 & 72.25 & 160.95 & 224.74 \\
8 & 0.1 & 72.50 & 71.00 & 71.75 & 126.90 & 189.34 \\
8 & 0.2 & 72.33 & 70.33 & 71.33 & 100.70 & 166.82 \\
10 & 0.05 & 72.50 & 73.00 & 72.75 & 161.22 & 225.80 \\
10 & 0.1 & 74.17 & 71.00 & 72.58 & 156.37 & 221.81 \\
10 & 0.2 & 72.83 & 70.67 & 71.75 & 127.64 & 190.52 \\
\bottomrule
\end{tabular*}
\end{table}

All other evaluation settings follow those in the main paper.
We repeat the evaluation on the six selected tasks for all compared configurations, measuring both success rate and latency.

Table~\ref{tab:app_robotwin_parameter_summary} reports the detailed results of the hyperparameter ablations shown in Fig.~\ref{fig:parameter_sensitivity}.

We detail the component ablations in Table~\ref{tab:mechanism_ablation} relative to the full samplers.
For SAG-Residual, the variant without residual checking accepts the entire window without correction, whereas the variant without Euler correction retains residual checking but commits the first failed draft unchanged.
For SAG-Coupling, constant velocity drafting replaces clean action drafting with a target velocity fixed at the window start and reevaluated at the reflected state after rejection.
Its latency increase includes the cost of reevaluating the velocity after rejection.
The variant without reference reuse retains clean action drafting but recomputes its reference at the reflected state after rejection.

\begin{table}[!htbp]
{
\setlength{\parindent}{1pc}
\section{Real-World Experimental Details}
\label{app:real_world_details}

For the real-world experiments in Fig.~\ref{fig:real_world}, we use D435 cameras to capture RGB images at a resolution of $640\times480$.
Actions are executed at 20~Hz, with \mbox{chunk length $H=32$} and \mbox{executed length $E=20$}.
The instructions for the four tasks are as follows:\par
\begin{enumerate}[label=\textbf{T\arabic*:},leftmargin=*,labelindent=0pt,labelsep=0.45em,align=left,itemsep=1pt,topsep=2pt,parsep=0pt,partopsep=0pt,before=\raggedright]
    \item \textbf{Banana pick-and-place.} Place a banana in a blue plastic box.
    \item \textbf{Block stacking.} Stack a blue block on a green block.
    \item \textbf{Toy storage.} Place a toy inside a drawer, then close the drawer.
    \item \textbf{Shuttlecock stacking.} Stack three shuttlecocks.
\end{enumerate}

Table~\ref{tab:app_real_world_sampler_profile} summarizes the sampler statistics for both SAG variants on the four real-world tasks, including integration steps advanced per full window, forward calls per action chunk, and target state evaluations per action chunk.
Both variants use $K=10$ and $M=4$, with $\delta=0.1$ for SAG-Residual and $\lambda=0.1$ for SAG-Coupling.
\par}
\vspace{6pt}

    \centering
    \captionsetup{skip=4pt}
    \caption{Sampler statistics on real-world tasks. Steps denote the mean number of integration steps advanced per full window of planned length $M=4$, including the correction step after rejection. Calls count action DiT forward passes, including initialization, with each batched pass counted once. States sum the batch sizes across calls. Calls and states are averaged per action chunk over inputs and random streams. Average rows report the unweighted mean across the four tasks.}
    \label{tab:app_real_world_sampler_profile}
    \small
    \setlength{\tabcolsep}{2pt}
    \begin{tabular*}{\columnwidth}{@{\extracolsep{\fill}}llrrr@{}}
        \toprule
        Task & Method & \shortstack{Steps /\\window} & \shortstack{Calls /\\chunk} & \shortstack{States /\\chunk} \\
        \midrule
        Banana & SAG-Residual & 4.00 & 4.11 & 11.11 \\
               & SAG-Coupling & 3.44 & 4.33 & 12.42 \\
        \addlinespace
        Blocks & SAG-Residual & 4.00 & 4.22 & 11.22 \\
               & SAG-Coupling & 3.41 & 4.64 & 12.75 \\
        \addlinespace
        Drawer & SAG-Residual & 4.00 & 4.00 & 11.00 \\
               & SAG-Coupling & 3.38 & 4.31 & 12.64 \\
        \addlinespace
        Shuttlecocks & SAG-Residual & 4.00 & 4.00 & 11.00 \\
                     & SAG-Coupling & 3.43 & 4.42 & 13.33 \\
        \midrule
        Average & SAG-Residual & 4.00 & 4.08 & 11.08 \\
                & SAG-Coupling & 3.42 & 4.42 & 12.78 \\
        \bottomrule
    \end{tabular*}
\end{table}

\end{document}